\pdfoutput=1

\documentclass[11pt]{article}

\usepackage[preprint]{acl}

\usepackage{times}
\usepackage{latexsym}

\usepackage[T1]{fontenc}

\usepackage[utf8]{inputenc}

\usepackage{microtype}

\usepackage{inconsolata}

\usepackage{graphicx}
\usepackage{subfigure}
\usepackage{booktabs} %

\usepackage{rotating}
\usepackage{adjustbox}

\usepackage{amsmath}
\usepackage{amssymb}
\usepackage{mathtools}
\usepackage{amsthm}
\usepackage{pifont}
\newcommand{\cmark}{\ding{51}}
\newcommand{\xmark}{\ding{55}}

\usepackage[most]{tcolorbox}
\usepackage{listings}

\usepackage[dvipsnames,table,xcdraw]{xcolor}
\usepackage{ulem}

\usepackage{hyperref}

\usepackage{longtable}

\usepackage{booktabs} 
\usepackage{setspace}

\usepackage{algorithm}
\usepackage{algpseudocode}

\usepackage{caption}

\usepackage{bm}
\usepackage{paralist}
\usepackage{array}
\usepackage{tabularx}
\usepackage{ragged2e}
\usepackage{makecell}
\usepackage{tikz}
\usepackage{pgfplots}
\usepackage{wrapfig}
\usetikzlibrary{pgfplots.groupplots, matrix}
\pgfplotsset{compat=1.18}
\usepackage{graphicx}
\usepackage{multirow}
\usepackage{stackengine}
\usepackage{colortbl} 
\usepackage{anyfontsize}
\usepackage{soul}
\usepackage{listings}
\usepackage{CJKutf8}

\DeclareRobustCommand{\Qwenemoji}{\includegraphics[height=0.9em]{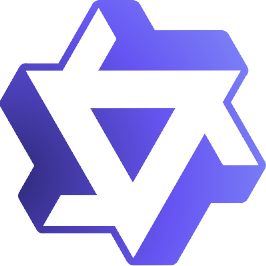}}
\DeclareRobustCommand{\Googleemoji}{\includegraphics[height=0.9em]{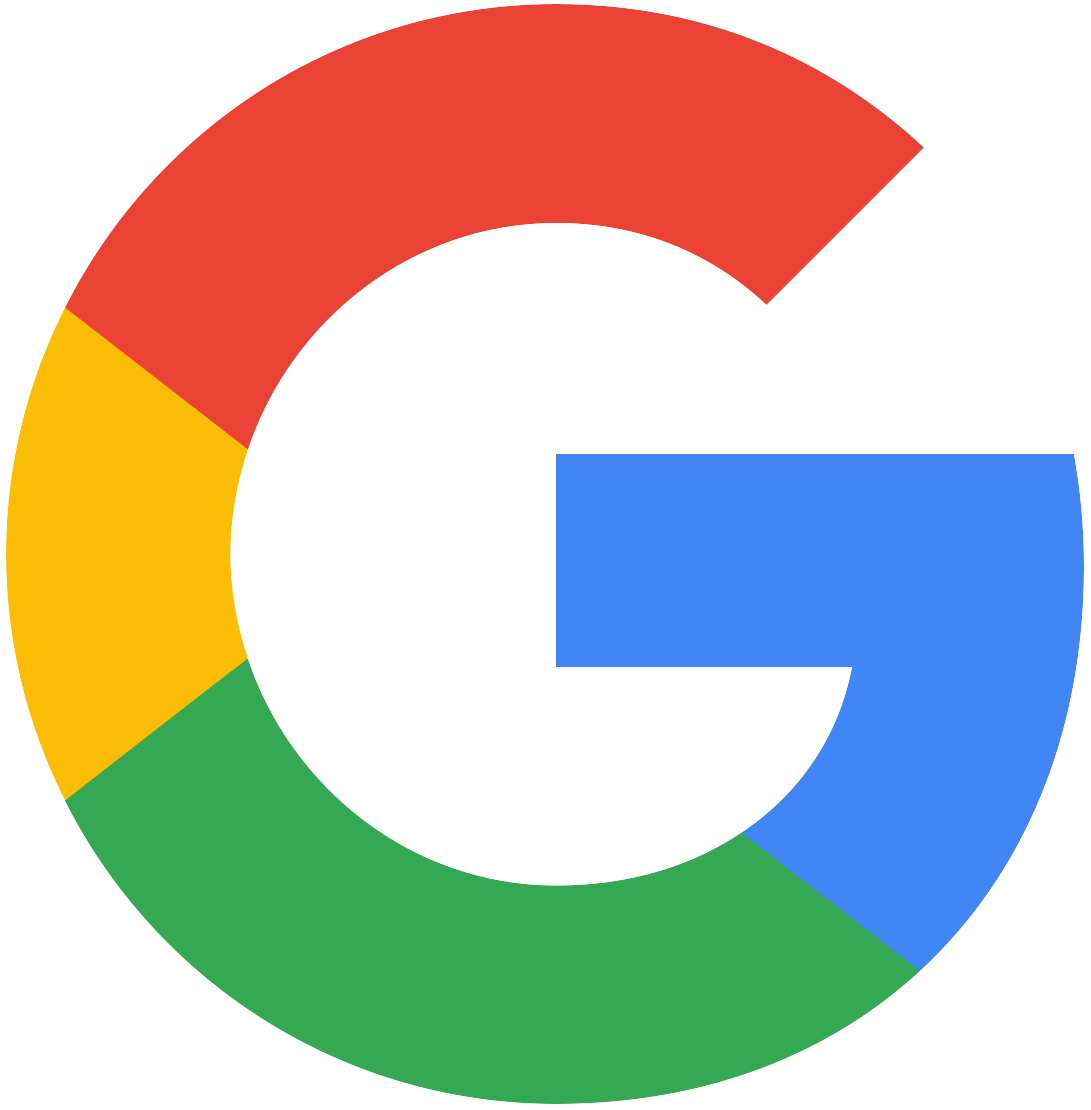}}
\DeclareRobustCommand{\Openaiemoji}{\includegraphics[height=0.8em]{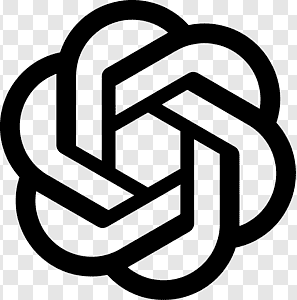}}
\DeclareRobustCommand{\pixtralemoji}{\includegraphics[height=0.9em]{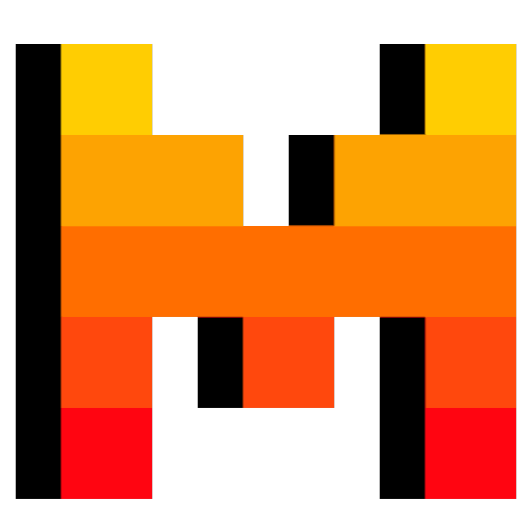}}

\DeclareRobustCommand{\ghemoji}{\includegraphics[height=0.8em]{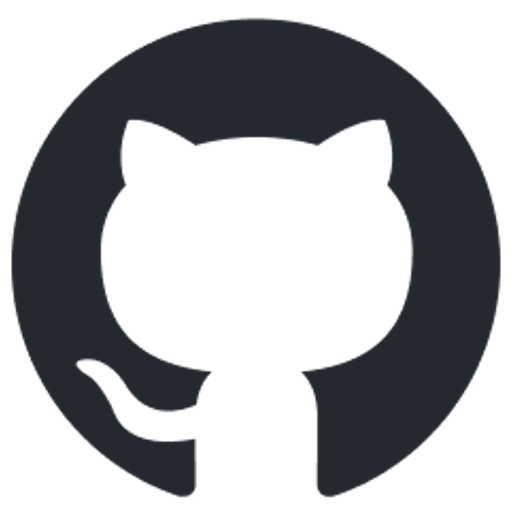}}

\definecolor{TaskBG}{HTML}{EFE6FF}        
\definecolor{StateBG}{HTML}{F5F5F7}       
\definecolor{ExpertBG}{HTML}{EAF7EA}      
\definecolor{IWMBG}{HTML}{FDECF3}         
\definecolor{SRBG}{HTML}{E6F2FF}          
\definecolor{GemmaBlueHex}{HTML}{1f77b4}
\definecolor{QwenGreenHex}{HTML}{2ca02c}
\definecolor{MistralOrangeHex}{HTML}{ff7f0e}

\newtcolorbox{trainingexample}[2][]{%
  enhanced, breakable, colframe=black!12, colback=white, boxrule=2.5pt,
  arc=2pt, left=0pt, right=0pt, top=0pt, bottom=0pt,
  title={#2}, fonttitle=\bfseries, coltitle=black, fontupper=\footnotesize, #1}
\newcommand{\EXrow}[3]{\rowcolor{#1}\textbf{#2} & #3\\}

\DeclareRobustCommand{\hlblue}[1]{{\sethlcolor{SRBG}\hl{#1}}}

\definecolor{codegreen}{rgb}{0,0.6,0}
\definecolor{codegray}{rgb}{0.5,0.5,0.5}

\definecolor{backcolour}{RGB}{245,248,250}
\definecolor{emph}{RGB}{166,88,53}
\definecolor{nightblue}{RGB}{9,49,105}
\definecolor{keywords}{RGB}{207,33,46}
\definecolor{lightpurple}{RGB}{130,81,223}

\lstdefinestyle{mystyle}{
    backgroundcolor=\color{backcolour},   
    commentstyle=\color{codegreen},
    keywordstyle=\color{keywords},
    stringstyle=\color{nightblue},
    basicstyle=\fontsize{7}{8}\ttfamily,
    breakatwhitespace=true,         
    breaklines=true,                 
    captionpos=b,                    
    keepspaces=true,                 
    numberstyle=\tiny\color{codegray},
    numbersep=2pt,                  
    showspaces=false,                
    showstringspaces=false,
    showtabs=false,                  
    tabsize=2,
    emph={dspy},
    emphstyle={\color{lightpurple}},
    linewidth=1\columnwidth,
    frame=tb,    
    xrightmargin=0pt,
    xleftmargin=0.23cm,
    numbers=left,
    aboveskip=0.2cm,
    belowskip=0.1cm,
}

\def\cquad{\hskip0.8em\relax}

\title{Adaptive Text Anonymization: \\
Learning Privacy-Utility Trade-offs via Prompt Optimization}

\author{Gabriel Loiseau$^{1,2}$ \cquad Damien Sileo$^{2}$ \cquad Damien Riquet$^{1}$ \cquad Maxime Meyer$^{1}$ \cquad Marc Tommasi$^{2}$ \\ $^{1}$Hornetsecurity, Hem, France \\ $^2$Univ. Lille, Inria, CNRS, Centrale Lille, UMR 9189 - CRIStAL, F-59000 Lille, France \\
\texttt{gabriel.loiseau@inria.fr}}

\begin{document}
\maketitle
\begin{abstract}
Anonymizing textual documents is a highly context-sensitive problem: the appropriate balance between privacy protection and utility preservation varies with the data domain, privacy objectives, and downstream application. However, existing anonymization methods rely on static, manually designed strategies that lack the flexibility to adjust to diverse requirements and often fail to generalize across domains. We introduce adaptive text anonymization, a new task formulation in which anonymization strategies are automatically adapted to specific privacy–utility requirements. We propose a framework for task-specific prompt optimization that automatically constructs anonymization instructions for language models, enabling adaptation to different privacy goals, domains, and downstream usage patterns. To evaluate our approach, we present a benchmark spanning five datasets with diverse domains, privacy constraints, and utility objectives. Across all evaluated settings, our framework consistently achieves a better privacy–utility trade-off than existing baselines, while remaining computationally efficient and effective on open-source language models, with performance comparable to larger closed-source models. Additionally, we show that our method can discover novel anonymization strategies that explore different points along the privacy–utility trade-off frontier.\footnote{Our data, code, and optimized prompts are available at \ghemoji{}~\url{https://github.com/gabrielloiseau/adaptive-text-anonymization}}
\end{abstract}
\section{Introduction}
Text anonymization has emerged as a fundamental technique for enabling the sharing and analysis of sensitive textual data while protecting individual privacy. 
Traditionally, privacy protection has been grounded in formal guarantees, most notably differential privacy \cite{dwork2006dp}, which provides strong, distribution-independent assurances against information leakage. However, applying such guaranties to free-form text remains notoriously difficult: defining neighboring datasets, calibrating noise, and preserving semantic utility are all unresolved challenges in practice \cite{mattern-etal-2022-limits, meisenbacher-etal-2025-impact, çano2025differentiallyprivatetextgenerationdegrades}. As a result, text anonymization is often conducted under a weaker but more operational notion of privacy, where an adversary with access to auxiliary background knowledge attempts to infer sensitive attributes from the released text.

Within this adversarial framing, large language models (LLMs) play a dual role. On the one hand, they constitute a powerful and realistic attacker model: LLMs encode broad world knowledge, can exploit subtle contextual cues, and have been shown to strategically re-identify sensitive information even after sophisticated anonymization \cite{staab2024beyond}. On the other hand, the same capabilities make LLMs effective anonymizers. When properly instructed, LLMs can perform nuanced, context-aware obfuscation of sensitive attributes, often outperforming rule-based or static methods \cite{staab2025anonymizers, yang-etal-2025-robust}. Recent work has demonstrated that collaborative settings involving an “attacker” LLM and an “anonymizer” LLM can further strengthen privacy, with multi-hop refinement yielding substantial improvements in anonymization quality \cite{yang-etal-2025-robust, brahem:hal-04684512}.


Nonetheless, successful anonymization is inherently context-dependent and must be evaluated across multiple dimensions. The effectiveness of anonymization strategies depends primarily on two key factors: the capabilities and resources of potential adversaries, and the specific notion of utility that must be preserved. These factors interact to create distinct anonymization requirements across different use cases. For instance, anonymizing a medical report containing sensitive patient information demands a fundamentally different approach than anonymizing casual online comments. Medical records may require protection against attackers capable of infering sensitive patient attributes, while preserving precise clinical terminology and diagnostic relationships essential for medical research. In contrast, online comments may face different threat models, such as inference attacks based on writing style or demographic cues. The diversity of these scenarios underscores that no single anonymization strategy can adequately address all privacy-utility requirements \cite{loiseau-etal-2025-tau}, necessitating adaptive approaches that can be tailored to specific adversarial assumptions and utility constraints \cite{pasch-cha-2025-balancing}. In practice, selecting a particular privacy–utility trade-off ultimately reflects an explicit risk assessment, in which acceptable levels of disclosure risk are weighed against task-specific utility requirements.

Despite these advances, modern LLM-based anonymization pipelines suffer from three key limitations that hinder their practical deployment. First, they adopt a \textit{fixed trade-off paradigm}, in which a single anonymization strategy is manually designed for each specific scenario, preventing flexible adaptation to new privacy-utility requirements. Second, they rely on \textit{manual prompt engineering}, requiring practitioners to iteratively design and refine instructions through trial and error—a process that is subjective, labor-intensive, and often yields suboptimal results. Third, they typically depend on \textit{proprietary closed-source models} to achieve effective anonymization, creating a fundamental contradiction: processing sensitive data through external API providers inherently conflicts with privacy objectives. These limitations combine to produce anonymization systems that are brittle across domains, costly to adapt to new contexts, and unsuitable for real-world deployment where sensitive content must remain under local control.

To address these challenges, we introduce \textit{adaptive text anonymization}, a framework that automatically learns domain- and task-specific anonymization strategies through prompt optimization. We formulate anonymization as a multi-objective optimization problem in which privacy and utility requirements are explicitly specified, targeting automatic adaptation. Our framework employs evolutionary prompt optimization and can operate on medium-sized language models, achieving performance comparable to proprietary API-based solutions while preserving full data privacy. Our contributions are as follows:
\begin{itemize}
\setlength\itemsep{-0.4em}
    \item We propose an adaptive anonymization framework that automatically discovers effective anonymization instructions through agentic collaboration and prompt optimization, enabling open-source models to navigate diverse privacy–utility trade-offs without manual prompt design.
    \item We introduce a unified multi-task evaluation setup spanning five anonymization tasks across distinct domains, privacy objectives, and utility constraints, providing a comprehensive evaluation suite for adaptive text anonymization.
    \item We present extensive empirical results demonstrating that our approach achieves state-of-the-art performance with open-source models and outperforms traditional anonymization methods both on their target task and on new domains.
    \item We show that our method enables explicit exploration of the privacy–utility trade-off space, discovering multiple anonymization strategies that favor different operating points and allowing practitioners to select models aligned with their deployment requirements.
\end{itemize}

\begin{figure*}[t]
  \centering
  \includegraphics[width=0.81\textwidth]{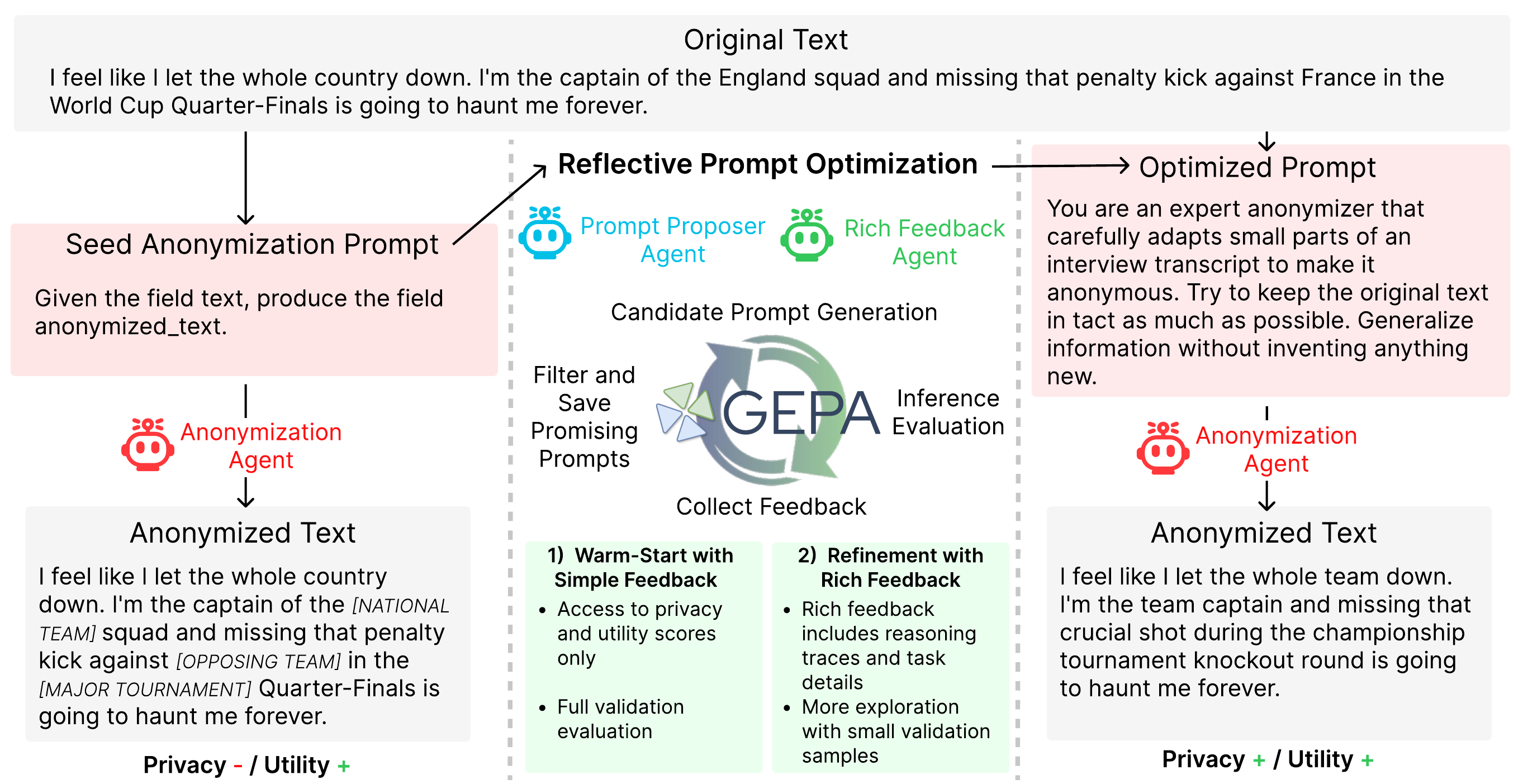}
  \caption{Overview of our approach. We perform reflective prompt optimization using the GEPA algorithm \cite{agrawal2025gepareflectivepromptevolution}. Our method adapts a base seed prompt into an optimized prompt defining the privacy and utility task requirements. The optimization operates in a strict fixed budget environment while learning sufficiently strong patterns to adapt to the anonymization objective.}
  \label{fig:framework_architecture}
  \vspace{-1em}
\end{figure*}

\section{Related Work}
\paragraph{Text Anonymization.}
Text anonymization seeks to enable the sharing of sensitive textual data by removing or obfuscating personal identifiers while preserving document utility. Traditional approaches predominantly rely on sequence labeling models trained on manually annotated datasets to detect and mask predefined categories of sensitive entities~\cite{deusser2025survey, hathurusinghe-etal-2021-privacy, francopoulo2020anonymization, lison-etal-2021-anonymisation}. While effective in structured privacy settings, these methods often overlook the impact of anonymization on downstream utility or restrict evaluation to surface-level text quality metrics~\cite{yermilov-etal-2023-privacy, staab2025anonymizers}. This limitation is further compounded by the lack of comprehensive benchmarks that systematically evaluate privacy–utility trade-offs across diverse domains, threat models, and task requirements~\cite{pasch-cha-2025-balancing, loiseau-etal-2025-tau}. 

\paragraph{Anonymization via LLM Pipelines.}
The advent of large language models has fundamentally reshaped text anonymization, establishing LLMs as both powerful privacy adversaries and potential defenders. Recent studies show that LLMs can successfully re-identify sensitive attributes in anonymized text~\cite{staab2024beyond, patsakis2023man}, exposing the limitations of traditional anonymization pipelines. In response, several works have explored the use of LLMs themselves as anonymization agents. \citet{staab2025anonymizers} introduced adversarial LLM collaboration, using a simulated attacker to guide anonymization decisions. Subsequent work has extended this paradigm by injecting synthetic information to mislead attackers while preserving utility~\cite{frikha2024incognitext}, or by incorporating multi-hop refinement among attacker, utility evaluator, and anonymizer LLMs to improve robustness~\cite{yang-etal-2025-robust}. Despite their effectiveness, these approaches rely heavily on manually designed instructions and domain-specific prompt engineering~\cite{dou-etal-2024-reducing, brahem:hal-04684512}, limiting their scalability and adaptability. Moreover, most methods depend on closed, API-based models (e.g., GPT-4/5 \cite{deusser2025survey}), raising practical and ethical concerns when processing sensitive data in deployment settings that require local control.

\paragraph{Prompt Optimization.} 
Automated prompt optimization has emerged as a promising alternative to manual prompt engineering, which is often inefficient, subjective, and difficult to generalize across tasks~\cite{yang2024large, 10.5555/3600270.3602070}. Recent work has explored search-based techniques~\cite{zhou2023large, opsahl-ong-etal-2024-optimizing} and evolutionary algorithms that iteratively refine prompts through mutation and selection~\cite{agrawal2025gepareflectivepromptevolution, chen2024instructzero}. These methods have demonstrated strong performance across a variety of NLP tasks, but, to the best of our knowledge, have not yet been applied to privacy-sensitive settings such as text anonymization. In particular, the potential of prompt optimization to automatically generate task-specific anonymization instructions aligned with explicit privacy objectives and utility constraints remains largely unexplored. Our work bridges this gap by leveraging evolutionary prompt optimization to enable adaptive, context-aware text anonymization without manual prompt design.

\section{Our Approach}
We introduce an agentic optimization framework for LLM-based text anonymization tasks. Our approach conceptualizes anonymization as a text generation task that can be learned and refined through natural language instruction and self-reflection. By framing the problem as prompt optimization, our framework enables a local anonymization agent to adapt its behavior to specific privacy objectives, data domains and downstream utility requirements without requiring manual prompt engineering or model retraining. This section formalizes the text anonymization problem, presents our two-stage optimization pipeline, and details the mechanisms enabling automatic adaptation across diverse privacy-utility trade-offs.

\subsection{Problem Formulation}
\label{sec:problem_formulation}
We view text anonymization as an adaptive, task-conditioned rewriting problem rather than a fixed set of redaction rules. In many practical deployments, the anonymizer must be configured \textit{before} processing data: (i) a \textbf{privacy specification} describes what must be protected (e.g., PII, quasi-identifiers, writing style, organization-specific secrets), and (ii) a \textbf{utility specification} describes what must be preserved to remain compatible with downstream use (e.g., clinical meaning, intent labels, formatting). This formulation aligns with regulations such as GDPR’s emphasis on specified purposes for data processing \cite{GDPRArticle5}. We call this setting \textit{adaptive task anonymization}.
Formally, an anonymization task is defined by a pair $(p,u)$, where $p$ is the privacy objective and $u$ is the utility objective. Given an input text $x$, the goal is to produce an anonymized text $\tilde{x}$ that satisfies $p$ while preserving the aspects required by $u$. These objectives are provided to the model as inputs rather than as fixed criteria, the same raw text may admit multiple acceptable anonymizations, reflecting different trade-offs aligned with distinct downstream applications.

Rather than retraining models for each domain, we represent the anonymizer as an LLM guided by a natural-language instruction prompt $\Pi$. The key challenge is to obtain an instruction $\Pi$ that is well-matched to the task $(p,u)$. Our contribution is to \textit{learn} $\Pi$ automatically from the task specification (i.e., its evaluation procedure) and empirical feedback. We start from a single universal \textit{seed prompt} $\Pi_0$ shared across all tasks and evolve it into a task-adapted prompt. This keeps the anonymization model unchanged and shifts adaptation to prompt optimization.

\subsection{Framework Overview}
\begin{algorithm}[t]
\scriptsize
\caption*{\scriptsize \textbf{Algorithm 1} Two-Phase GEPA for Text Anonymization}
\label{alg:cap}
\begin{algorithmic}[1]
\Require Dataset $D$, base feedback metric $\mu$, budget $B$, local model $m$, feedback creation model $\mathcal{M}$
\Require Base prompt $\Pi_0$, early-stop patience $n$, sampling ratio $\alpha$
\State \textbf{Stage 1: Initialization}
\State Split $D$ into $D_{\text{train}}$ and $D_{\text{valid}}$
\State Initialize prompt pool $P \gets \{\Pi^m_0\}$
\State \textbf{Stage 2: Warm-Start with Basic Feedback}
\While{$B > 0$ \textbf{and} no improvement for $< n$ iterations}
\State $\Pi^m_{\text{new}} \gets \text{GEPA}(P, D_{\text{train}}, D_{\text{valid}}, \mu, m)$
\State $P \gets P \cup \{\Pi^m_{\text{new}}\}$; update $B$
\EndWhile
\State \textbf{Stage 3: Refinement with Rich Feedback}
\If{$B > 0$}
\State $\mu_{\text{rich}} \gets \mathcal{M}(\mu)$
\While{$B > 0$}
\State $D'_{\text{valid}} \gets \text{Sample}(D_{\text{valid}}, \alpha)$
\State $\Pi^m_{\text{new}} \gets \text{GEPA}(P, D_{\text{train}}, D'_{\text{valid}}, \mu_{\text{rich}}, m)$
\State $P \gets P \cup \{\Pi^m_{\text{new}}\}$; update $B$
\EndWhile
\EndIf
\State \Return $\arg\max_{\Pi^m \in P} \text{Score}(\Pi^m, D_{\text{valid}})$
\end{algorithmic}
\end{algorithm}

Algorithm~1 operationalizes adaptive task anonymization as prompt evolution and optimization. We explore anonymization instructions under a fixed computational budget $B$, measured in LLM forward passes. The framework operates entirely using a locally deployable language model $m$, which serves two roles: (i) an \textit{anonymization agent} that applies candidate instructions to produce anonymized text, and (ii) a \textit{proposer agent} that generates new instruction variants during optimization.

Our approach builds upon \textbf{GEPA}~\cite{agrawal2025gepareflectivepromptevolution}, an evolutionary prompt optimization algorithm that treats prompts as individuals in a population $P$, evolved through selection, mutation, and evaluation. Candidate prompts are retained via Pareto-based selection over privacy and utility, encouraging diverse trade-off strategies. GEPA employs \textit{reflective mutation}: the proposer analyzes execution traces and feedback to propose targeted modifications.

\subsection{Stage 1: Initialization}
The optimization begins with partitioning the dataset $D$ into training and validation splits, denoted $D_{\text{train}}$ and $D_{\text{valid}}$. The training set is used to generate anonymized outputs during candidate evaluation, while the validation set defines the fitness landscape used to score candidate instructions. We initialize the prompt pool $P$ with a single candidate $\Pi^m_0$, which pairs a generic seed instruction $\Pi_0$ (\textit{"Given the field `text`, produce the field `anonymized\_text`."}) with the local model $m$. This seed provides a minimal anonymization behavior from which more specialized strategies can evolve. During early optimization, candidate prompts are evaluated using a \textit{base feedback function} $\mu$, which serves as a learnable proxy for the task specification $(p,u)$ defined in Section~\ref{sec:problem_formulation}. The function $\mu$ is typically defined as a scalar aggregation of privacy and utility evaluation metrics (i.e., their average) without additional detail, which provides a coarse signal to guide exploration.

\subsection{Stage 2: Warm-Start with Basic Feedback}
Stage~2 applies standard GEPA to explore the instruction space using base feedback $\mu$. Each iteration: (1) selects diverse, high-performing prompts from the pool $P$ using Pareto-based ranking over privacy and utility objectives; (2) The proposer agent analyzes execution traces and feedback associated with a selected prompt on a minibatch of texts in $D_{\text{train}}$ and proposes targeted instruction modifications, yielding a new candidate $\Pi^m_{\text{new}}$. And (3) The new prompt is applied to the same minibatch of $D_{\text{train}}$. If an improvement is observed on this minibatch, it gets evaluated on $D_{\text{valid}}$ using $\mu$, and incorporated into $P$, with dominated candidates removed via Pareto pruning. Stage~2 terminates when either the budget $B$ is exhausted or validation performance plateaus for $n$ consecutive iterations. This early-stopping criterion preserves budget for later refinement.


\subsection{Stage 3: Refinement with Rich Feedback}
If computational budget remains, the algorithm transitions to a refinement phase that aims to escape local optima and achieving finer-grained control over the privacy–utility trade-off. This phase extends standard GEPA with two novel mechanisms.
\paragraph{Rich Feedback Generation.}
A \textit{rich feedback function} $\mu_{\text{rich}}$ is derived from the task-specific metric definitions by decomposing the aggregate score $\mu$. In addition to scalar values, $\mu_{\text{rich}}$ may include natural-language explanations or evaluator reasoning traces that provide interpretable, structured feedback to the proposer agent. This feedback function is generated once per task by a separate LLM (referred to as the rich feedback agent in Figure~\ref{fig:framework_architecture}), ensuring consistency and avoiding manual, subjective feedback design\footnote{In practice, we assign the rich feedback agent a code-generation task using the implementation of $\mu$ in order to obtain the corresponding implementation of $\mu_{rich}$. More information in Appendix~\ref{apx:implementation_details}.}.
\paragraph{Adaptive Validation Sampling.}
To improve exploration under the remaining budget, we evaluate candidate prompts over sampled validation subsets $D'_{\text{valid}} \subset D_{\text{valid}}$. Sampling follows a round-robin strategy prioritizing under-evaluated examples, balancing computational efficiency with coverage diversity. This reduces the budget consumption during evaluation while mitigating overfitting. Final model selection is performed using the full validation set to ensure fair comparison.

With these mechanisms, GEPA continues to evolve the prompt pool until the budget is exhausted. The richer feedback enables the proposer agent to perform larger, even more targeted behavioral updates using fewer evaluations. The framework ultimately returns the best-performing anonymization system $\Pi^*$ identified during optimization, corresponding to the highest validation score under the specified privacy–utility configuration.

\section{Experiments}

\paragraph{Evaluation Setup}
We evaluate our framework on five anonymization tasks spanning diverse domains, privacy threat models, and utility requirements. Each task poses distinct challenges in balancing privacy protection against information preservation, reflecting the context-dependent nature of real-world anonymization. The datasets are drawn from prior work~\cite{yang-etal-2025-robust, xin2025a, li-etal-2025-papillon}, and we retain their original evaluation protocols and implementations whenever available to ensure fair comparison. Additional dataset details are provided in Appendix~\ref{apx:task_details}.

\noindent
\textbf{DB-Bio}~\cite{yang-etal-2025-robust}. A collection of DBpedia biographies of notable individuals. 
\underline{Privacy}: A binary re-identification metric in which an attacker LLM attempts to recover the individual’s identity by proposing the top-3 candidate names from the anonymized biography. 
\underline{Utility}: Occupation classification accuracy measured on anonymized text.

\noindent
\textbf{SynthPAI}~\cite{yukhymenko2024synthetic}. Synthetic Reddit-style posts encoding demographic attributes (e.g., age, gender, location) through writing style and contextual cues. 
\underline{Privacy}: A binary inference metric indicating whether an attacker LLM can correctly predict target demographic attributes from the anonymized text. 
\underline{Utility}: ROUGE-1 F-measure assessing content preservation.

\noindent
\textbf{TAB}~\cite{pilan-etal-2022-text}. European court case documents containing personal information about involved parties and witnesses. 
\underline{Privacy}: Recall of correctly masked sensitive spans relative to gold-standard annotations. 
\underline{Utility}: Semantic similarity between original and anonymized documents.

\noindent
\textbf{PUPA}~\cite{zhao2024wildchat}. User prompts from ChatGPT interactions containing explicit personally identifiable information (PII), including names, addresses, phone numbers, and email addresses. 
\underline{Privacy}: PII leakage rate, defined as the fraction of sensitive entities remaining in the sanitized prompts. 
\underline{Utility}: An LLM-as-judge evaluation comparing response quality between original and anonymized prompts.

\noindent
\textbf{MedQA}~\cite{jin2021disease}. Clinical case descriptions drawn from USMLE-style medical examinations. 
\underline{Privacy}: Stylometric distance between original and sanitized texts, computed using LUAR embeddings \cite{rivera-soto-etal-2021-learning}, to reduce the risk of patient re-identification. 
\underline{Utility}: Accuracy of selecting the correct medical diagnosis from multiple-choice options.

\paragraph{Comparison Methods.}
To assess the effectiveness of our framework, we compare against five baseline methods that span traditional entity-based anonymization, adversarial LLM pipelines, and manual prompt engineering. Together, these baselines cover the dominant paradigms in contemporary text anonymization.

\begin{itemize}
    \setlength\itemsep{-0.5em}
    \item \textbf{OpenPII:} A supervised PII detection model fine-tuned on the OpenPII dataset\footnote{\url{https://hf.co/datasets/ai4privacy/open-pii-masking-500k-ai4privacy}}, using ModernBERT~\cite{warner-etal-2025-smarter}, representing entity-centric anonymization via explicit span detection.
    \item \textbf{DP-Prompt:} A document-level Local Differential Privacy (LDP) mechanism that builds on a paraphrasing model \citep{utpala-etal-2023-locally}. It privatizes text by clipping model logits and applying randomized perturbations to enforce a local DP guarantee. Due to the impact of noise on the utility harm, we set the privacy budget $\epsilon = 100$.
    \item \textbf{Adversarial Feedback (AF):} The method proposed by \citet{staab2025anonymizers}, using adversarial collaboration between two LLM agents: an attacker that attempts re-identification and an anonymizer that iteratively refines its outputs based on attacker feedback.
    \item \textbf{RUPTA:} The approach of \citet{yang-etal-2025-robust}, extending adversarial feedback with an additional utility judge. We evaluate RUPTA exclusively on \textsc{DB-Bio}, as it was designed for classification-oriented utility objectives.
    \item \textbf{Task-Specific Manual Prompting:} Expert-crafted prompts encoding privacy and utility requirements for each task, isolating the benefit of automatic prompt optimization.
\end{itemize}

\paragraph{Implementation Details.}
All implementations and optimizations in this paper are carried out using the DSPy~\cite{khattab2024dspy} library\footnote{Preliminary experiments confirm that the baseline methods are fully reproducible, showing no statistically significant differences between the original authors’ implementations and those using DSPy.}. We evaluate our framework using three open-source language models: \pixtralemoji{}~Mistral-Small-3.2-24B\footnote{\url{https://hf.co/mistralai/Mistral-Small-3.2-24B-Instruct-2506}}, \Googleemoji{}~Gemma-3-27B \cite{gemmateam2025gemma3technicalreport}, and \Qwenemoji{}~Qwen3-30B-A3B \cite{qwen3technicalreport}. Following GEPA experiment settings~\cite{agrawal2025gepareflectivepromptevolution} and to simulate realistic deployment with limited access to labeled sensitive data, we restrict training to a small, fixed set of 111 examples for training and 111 examples for validation per task ($|D_{\text{train}}| = |D_{\text{valid}}| = 111$), reserving all remaining examples exclusively for testing. We configure the evolutionary optimization with a maximum rollout budget of $B = 1500$ LLM forward passes, an early-stop patience of $n=5$ iterations and set the adaptive validation sampling ratio to $\alpha = 0.3$. This configuration balances exploration of the instruction space with computational efficiency across tasks and models. For evaluation, we report each benchmark’s primary native metric and adhere to the official evaluation implementations whenever available to ensure comparability with prior work. Privacy and utility metrics that require model-based evaluation are computed using \Googleemoji{}~Gemini-2.5-flash \cite{comanici2025gemini25pushingfrontier} as a strong and consistent evaluator backbone. As a reference closed-source model for LLM-based comparison baselines, we use \Openaiemoji{}~GPT-5-chat~\cite{gpt5systemcard} as the underlying model to reflect state-of-the-art performance and to faithfully reproduce original method behavior. Additional implementation details, including prompts, hyperparameters, and computational costs, are provided in Appendix~\ref{apx:implementation_details}. An ablation study analyzing the contribution of each component in our optimization approach is provided in Appendix~\ref{apx:ablation_study}.

\begin{table*}[htb]
    \vspace{-0.5em}
    \centering
    \vspace{-0.2em}
    \begin{adjustbox}{max width=0.9\textwidth}
    \small
    \begin{tabular}{ll|cc|>{\columncolor{IWMBG}}c >{\columncolor{SRBG}}c}
    \toprule
    \textbf{Benchmark} & \textbf{Model} & AF~\cite{staab2025anonymizers} & Task-Specific Prompt & \textbf{Seed Prompt} & \textbf{Optimized Prompt} \\
    \midrule
    \multicolumn{6}{c}{Privacy Score $\uparrow$ / Utility Score $\uparrow$}\\ \hline \midrule
    \multirow{3}{*}{DB-Bio} 
     & \pixtralemoji{}~Mistral-Small-3.2-24B & 59.7 / 98.1 & 58.8 / 100 & 54.5 / 98.9 & \textbf{61.2 / 100} \\
     & \Googleemoji{}~Gemma-3-27B-it & 59.1 / 94.2 & 47.1 / 100 & 67.6 / 99.1 & \textbf{77.6 / 100} \\
     & \Qwenemoji{}~Qwen3-30B-A3B & 58.8 / 97.2 & 53.2 / 100 & 64.0 / 100 & \textbf{65.5 / 100} \\ \midrule
    \multirow{3}{*}{SynthPAI}
     & \pixtralemoji{}~Mistral-Small-3.2-24B & 35.3 / 67.6 & 5.90 / 97.1 & 6.30 / \underline{99.2} & \underline{38.7} / 77.4 \\
     & \Googleemoji{}~Gemma-3-27B-it & \underline{40.7} / 60.2  & 23.5 / 70.4 & 36.0 / 62.9 & 36.0 / \underline{77.1} \\
     & \Qwenemoji{}~Qwen3-30B-A3B & \underline{31.3} / 56.0 & 7.32 / 93.7 & 12.6 / 92.4 & 22.5 / \underline{94.4} \\ \midrule
    \multirow{3}{*}{TAB}
     & \pixtralemoji{}~Mistral-Small-3.2-24B & 45.1 / 53.5 & 39.2 / \underline{55.5} & 20.8 / 54.3 & \underline{83.6} / 53.7 \\
     & \Googleemoji{}~Gemma-3-27B-it & 62.1 / 53.7 & 68.3 / 51.3 & 61.8 / \underline{55.9} & \underline{81.9} / 54.1 \\
     & \Qwenemoji{}~Qwen3-30B-A3B & 26.1 / 54.4 & 36.4 / 52.4 & 36.2 / 54.0 & \textbf{92.3 / 56.2} \\
    \midrule
    \multirow{3}{*}{PUPA}
     & \pixtralemoji{}~Mistral-Small-3.2-24B & 77.1 / 76.5 & 82.7 / 76.2 & 75.0 / 77.5 & \textbf{85.3 / 82.9} \\
     & \Googleemoji{}~Gemma-3-27B-it & 93.9 / 64.7 & 80.7 / 76.5 & 88.1 / 77.1 & \textbf{94.3 / 79.1} \\
     & \Qwenemoji{}~Qwen3-30B-A3B & 86.3 / 70.6 & 94.1 / 78.2 & 82.3 / 72.1 & \textbf{98.0 / 79.3} \\ \midrule
    \multirow{3}{*}{MedQA}
     & \pixtralemoji{}~Mistral-Small-3.2-24B & \underline{9.39} / 41.2 & 7.33 / 35.3 & 2.91 / \underline{54.9} & 8.06 / 53.2 \\
     & \Googleemoji{}~Gemma-3-27B-it & 11.7 / 29.4 & 10.7 / 35.3 & 12.0 / 46.8 & \textbf{14.3 / 56.8} \\
     & \Qwenemoji{}~Qwen3-30B-A3B & 6.53 / 29.4 & 8.25 / 41.2 & 3.52 / \underline{58.6} & \underline{24.6} / 45.9 \\ \midrule
    \bottomrule
    \end{tabular}
    \end{adjustbox}
    \caption{Results on open source models across all benchmark tasks. The best overall privacy–utility trade-off is shown in \textbf{bold}. Otherwise, the best score for each individual component is \underline{underlined}. Prompt optimization consistently yields stronger privacy-utility trade-offs than static anonymization baselines and task-specific prompts, often improving privacy substantially while preserving or even improving utility across diverse tasks and model families.}
    \label{tab:all_benchmarks}
    \vspace{-1em}
\end{table*}

\section{Experimental Results}
\subsection{Overall Performance}
We evaluate along two axes: (i) prompt optimization impact on open-source models relative to anonymization baselines applied to the same model, and (ii) competitiveness against GPT-5–based methods. We report test \emph{Privacy} and \emph{Utility} scores throughout (higher percentage is better).

Table~\ref{tab:all_benchmarks} summarizes results across all tasks and open-source models. The \textbf{Optimized Prompt} consistently improves privacy over the \textbf{Seed Prompt} and is frequently the strongest or best-balanced open-source approach. These gains are typically achieved with minimal utility loss, indicating that the optimizer discovers task-appropriate trade-offs rather than optimizing a single objective.

Privacy gains are largest on structurally amenable tasks such as \textsc{TAB} and \textsc{PUPA}, where optimization substantially improves privacy while preserving semantics or response quality. On \textsc{DB-Bio}, optimized prompts maintain near-perfect utility while improving privacy, reflecting the relatively permissive trade-offs of classification-based metrics. Although \textsc{SynthPAI} and \textsc{MedQA} present stronger privacy–utility tension, optimization still yields consistent improvements over seed prompts across models.

Models exhibit consistent, distinct behaviors. Mistral exhibits steep privacy gains, sometimes at the cost of utility; Gemma favors more conservative improvements building on the privacy-focused behavior established by the seed prompt; and Qwen is the most robust, frequently achieving high privacy and utility simultaneously. These patterns suggest that the optimizer adapts to model-specific inductive biases rather than converging to a single anonymization strategy.
Table~\ref{tab:rotated_results} compares one of our open-source configuration with GPT-5–based AF, task-specific prompting, RUPTA, OpenPII, and DP-Prompt. The latter two serve as baselines for utility-focused and privacy-focused methods, respectively\footnote{A comparison across different privacy budgets for DP-Prompt is provided in Appendix~\ref{apx:dp_comparison}}. Despite using a smaller model, optimized Qwen is competitive across all tasks: it matches GPT-5 on \textsc{MedQA}, achieves higher utility at comparable privacy on \textsc{PUPA}, and attains the strongest overall utility scores. This demonstrates that adaptive prompt optimization substantially narrows the gap between open- and closed-source anonymization pipelines. The benchmarks expose distinct privacy–utility landscapes. \textsc{DB-Bio} admits near-optimal solutions on both axes, while \textsc{SynthPAI} and \textsc{MedQA} enforce stricter trade-offs. Across these regimes, our framework consistently identifies effective strategies given the intrinsic constraints of each task.

\begin{table}[t]
    \centering
    \setlength{\tabcolsep}{3pt} 
    \renewcommand{\arraystretch}{1.2}
    
    \resizebox{\columnwidth}{!}{
    \begin{tabular}{lccccc}
    \textbf{Methods}
     & \textbf{DB-Bio} 
     & \textbf{SynthPAI}
     & \textbf{TAB}
     & \textbf{PUPA} 
     & \textbf{MedQA} \\
    \midrule
    OpenPII & 57.6/98.1 & 9.02/97.3 & 87.1/32.2 & 75.4/70.3 & 3.80/59.5 \\
    DP-Prompt & 82.1/13.2 & 95.2/1.49 & 99.2/37.4 & 84.2/49.1 & 51.6/9.82 \\
    \midrule
    RUPTA (\Openaiemoji{}~GPT-5) & 74.0/98.3 & - / - & - / - & - / - & - / - \\
    AF (\Openaiemoji{}~GPT-5) & 78.0/92.1 & 64.0/57.6 & 59.9/42.5 & 94.2/46.0 & 24.4/45.8 \\
    Prompt (\Openaiemoji{}~GPT-5) & 63.6/100 & 18.3/88.1 & 99.3/48.6 & 99.1/72.7 & 10.7/45.5 \\
    \midrule
    \rowcolor{SRBG} \Qwenemoji{} (Optimized) & 65.5/100 & 22.5/94.4 & 92.3/56.2 & 98.0/79.3 & 24.6/45.9 \\
    \bottomrule
    \end{tabular}
    }
    \caption{Performance of optimized Qwen3-30B-A3B compared to GPT-5 based methods, OpenPII and DP-Prompt. Models not adaptable on other tasks are marked with -.}
    \label{tab:rotated_results}
    \vspace{-1em}
    \end{table}

\subsection{Trade-off Discovery}
\begin{figure*}[htb]
    \centering
    \begin{tikzpicture}

    \colorlet{imitationLine}{black!85}
    \colorlet{imitationNode}{black!75}
    \colorlet{transparent}{white!100}
    \colorlet{IWMLine}{RubineRed!25}
    \colorlet{IWMNode}{RubineRed!75}
    \colorlet{SRLine}{RubineRed!35}
    \colorlet{SRNode}{RubineRed!95}

    \colorlet{GemmaLine}{GemmaBlueHex!25}
    \colorlet{GemmaNode}{GemmaBlueHex!95}
    \colorlet{QwenLine}{QwenGreenHex!25}
    \colorlet{QwenNode}{QwenGreenHex!95}
    \colorlet{MistralLine}{MistralOrangeHex!25}
    \colorlet{MistralNode}{MistralOrangeHex!95}
    
    \colorlet{AFNode}{violet}
    \colorlet{PromptNode}{teal}

        \begin{groupplot}[
            group style={
                group size=3 by 1,
                horizontal sep=0.8cm,
                xlabels at=edge bottom,
                ylabels at=edge left
            },
            width=6cm,
            height=6cm,
            grid=major,
            grid style={black!10},
            tick label style={font=\footnotesize},
            xlabel style={font=\scriptsize},
            ylabel style={font=\scriptsize},
        ]
    
        \nextgroupplot[
          ylabel={Utility Score (\%)},
          xmin=35, xmax=105, 
          ymin=41, ymax=57,  
          xtick={40,50,60,70,80,90,100},
          ytick={42, 44, 46, 48, 50, 52, 54, 56}, 
          legend style={legend to name=sharedlegend, legend columns=5, draw=none, font=\small},
        ]
        \addlegendimage{draw=MistralLine, thick, mark=square*, mark options={fill=MistralNode, draw=MistralNode}}
        \addlegendentry{Mistral-Small-3.2-24B}
        
        \addlegendimage{draw=GemmaLine, thick, mark=triangle*, mark options={fill=GemmaNode, draw=GemmaNode}}
        \addlegendentry{Gemma-3-27b-it}
        
        \addlegendimage{draw=QwenLine, thick, mark=*, mark options={fill=QwenNode, draw=QwenNode}}
        \addlegendentry{Qwen3-30B-A3B}
        
        \addlegendimage{draw=transparent, mark=star, mark options={fill=AFNode, draw=AFNode}}
        \addlegendentry{AF (GPT-5)}
        
        \addlegendimage{draw=transparent, mark=diamond*, mark options={fill=PromptNode, draw=PromptNode}}
        \addlegendentry{Prompt (GPT-5)}
        
        %
        \addplot[MistralLine, mark options={fill=MistralNode, draw=MistralNode}, mark=square*, mark size=1, thick] coordinates {(83.6, 53.7) (86.0, 51.54) (84.52, 51.54) (77.58, 51.76) (40.54, 53.36) (83.6, 53.7)};
        \addplot[only marks, MistralNode, mark options={fill=MistralNode, draw=MistralNode}, mark=square*, mark size=1] coordinates {(45.25,53.3) (84.52,51.54) (40.54, 53.36) (45.25, 53.3) (47.41, 53.1) (51.81, 53.02) (77.58, 51.76) (50.63, 53.02) (70.81, 53.11) (86.0, 51.54) (83.6, 53.7)};
        %
        \addplot[GemmaLine, mark options={fill=GemmaNode, draw=GemmaNode}, mark=triangle*, mark size=1.5, thick] coordinates {(81.9, 54.1) (73.78,52.37) (74.96,51.89) (79.82, 51.4) (95.06,51.11) (81.9, 54.1)};
        \addplot[only marks, GemmaNode,
              mark options={fill=GemmaNode, draw=GemmaNode},
              mark=triangle*, mark size=1.5
            ] coordinates {
              (84.82,51.63)
              (79.82,51.40)
              (79.06,52.20)
              (73.78,52.37)
              (95.06,51.11)
              (81.64,51.37)
              (74.96,51.89)
              (85.10,51.45)
              (77.89,52.21)
              (83.92,51.63)
              (81.9, 54.1)
            };
        %
        \addplot[QwenLine, mark options={fill=QwenNode, draw=QwenNode}, mark=*, mark size=1.5, thick] coordinates {(92.3, 56.2) (93.13, 54.43) (84.98, 53.76) (61.44, 53.77) (92.3, 56.2)};
        \addplot[only marks, QwenNode,
          mark options={fill=QwenNode, draw=QwenNode},
          mark=*, mark size=1.5
        ] coordinates {
          (84.82,54.56)
          (84.82,54.36)
          (86.59,54.96)
          (86.15,54.32)
          (93.13,54.43)
          (86.15,54.80)
          (61.44,53.77)
          (86.00,54.28)
          (84.98,53.76)
          (86.15,54.33)
          (92.3, 56.2)
        };
        
        \addplot[only marks, AFNode, mark options={fill=AFNode, draw=AFNode}, mark=star, mark size=2.5] coordinates {(59.9, 42.5)};
        
        \addplot[only marks, PromptNode, mark options={fill=PromptNode, draw=PromptNode}, mark=diamond*, mark size=2.5] coordinates {(99.3, 48.6)};

        \addplot[imitationLine, mark options={fill=imitationNode, draw=imitationNode},  mark=*, mark size=0, dashed] coordinates {(92.3, 56.2) (93.13,54.43) (95.06,51.11) (99.3, 48.6)};

        \nextgroupplot[
          xmin=0, xmax=70,
          ymin=30, ymax=102,
          xtick={10,20,30,40,50,60,70}, 
          ytick={40,50,60,70,80,90},
        ]

        \addplot[MistralLine, mark options={fill=MistralNode, draw=MistralNode}, mark=square*, mark size=1, thick] coordinates {(38.18,60.56) (41.82,65.4) (38.7, 77.4) (32.73,81.45) (21.82,90.32) (30.91,64.35) (38.18,60.56)};
        \addplot[only marks, MistralNode, mark options={fill=MistralNode, draw=MistralNode}, mark=square*, mark size=1] coordinates {
          (27.27,74.91)
          (41.82,65.40)
          (32.73,76.44)
          (30.91,64.35)
          (36.36,65.71)
          (32.73,81.45)
          (27.27,79.30)
          (21.82,90.32)
          (38.18,60.56)
          (29.09,81.30)
          (38.7, 77.4)
        };

        \addplot[GemmaLine, mark options={fill=GemmaNode, draw=GemmaNode}, mark=triangle*, mark size=1.5, thick] coordinates {(32.73,77.76) (29.09,69.49) (34.55,70.1) (38.18,72.67) (38.18,76.91) (32.73,77.76)};
        \addplot[only marks, GemmaNode,
              mark options={fill=GemmaNode, draw=GemmaNode},
              mark=triangle*, mark size=1.5
            ] coordinates {
              (32.73,75.78)
              (32.73,77.76)
              (32.73,76.19)
              (29.09,69.49)
              (38.18,74.41)
              (34.55,74.23)
              (38.18,76.91)
              (38.18,72.67)
              (34.55,70.10)
              (30.91,70.10)
              (36.0, 77.1)
            };

        \addplot[QwenLine, mark options={fill=QwenNode, draw=QwenNode}, mark=*, mark size=1.5, thick]
        coordinates {
          (7.27,99.15)
          (38.18,34.66)
          (49.09,34.87)
          (45.45,47.91)
          (34.55,77.11)
          (22.5,94.4)
          (7.27,99.15)
        };
        \addplot[only marks, QwenNode,
          mark options={fill=QwenNode, draw=QwenNode},
          mark=*, mark size=1.5
        ] coordinates {
          (12.73,93.38)
          (45.45,47.91)
          (21.82,83.41)
          (21.82,84.81)
          (34.55,77.11)
          (7.27,99.15)
          (38.18,34.66)
          (38.18,44.56)
          (49.09,34.87)
          (14.55,96.75)
          (22.5, 94.4)
        };

        \addplot[only marks, AFNode, mark options={fill=AFNode, draw=AFNode}, mark=star, mark size=2.5] coordinates {(64.0, 57.6)};
        \addplot[only marks, PromptNode, mark options={fill=PromptNode, draw=PromptNode}, mark=diamond*, mark size=2.5] coordinates {(18.3, 88.1)};

        \addplot[imitationLine, mark options={fill=imitationNode, draw=imitationNode},  mark=*, mark size=0, thick, dashed] coordinates {(7.27,99.15) (14.55,96.75) (22.5,94.4) (32.73,81.45) (38.18,76.91) (41.82,65.4) (64.0, 57.6)};

        \nextgroupplot[
          xmin=1.5, xmax=28,
          ymin=45, ymax=67,
          xtick={5,10,15,20,25},
          ytick={47.5,50,52.5,55.0,57.5,60.0,62.5,65.0}
        ]
        
        \addplot[MistralLine, mark options={fill=MistralNode, draw=MistralNode}, mark=square*, mark size=1, thick] coordinates {(3.04, 61.82) (3.26, 54.55) (8.22, 47.27) (14.58, 49.27) (8.63, 60.0) (3.04, 61.82)};
        \addplot[only marks, MistralNode, mark options={fill=MistralNode, draw=MistralNode}, mark=square*, mark size=1] coordinates {
          (3.26, 54.55) (3.04, 61.82) (4.29, 60.0) (6.52, 56.36) 
          (5.92, 54.55) (14.58, 49.27) (8.63, 60.0) (8.22, 47.27) 
          (7.47, 54.55) (7.74, 52.73)
          (8.06, 53.2)
        };
        
        \addplot[GemmaLine, mark options={fill=GemmaNode, draw=GemmaNode}, mark=triangle*, mark size=1.5, thick] coordinates {(5.87, 56.36) (9.46, 56.36) (14.3, 56.8) (10.23, 65.45) (6.09, 61.82) (5.87, 56.36)};
        \addplot[only marks, GemmaNode,
              mark options={fill=GemmaNode, draw=GemmaNode},
              mark=triangle*, mark size=1.5
            ] coordinates {
              (6.48, 56.36) (5.87, 56.36) (6.09, 61.82) (7.09, 60.0) 
              (7.45, 58.18) (9.46, 56.36) (8.35, 63.64) (6.8, 56.36) 
              (10.23, 65.45) (9.37, 58.18)
              (14.3, 56.8)
            };
            
        \addplot[QwenLine, mark options={fill=QwenNode, draw=QwenNode}, mark=*, mark size=1.5, thick] coordinates {(10.51, 60.0) (8.03, 58.18) (12.06, 50.91) (16.72, 47.27) (24.6, 45.9) (15.59, 56.36) (12.47, 60.0) (10.51, 60.0)};
        \addplot[only marks, QwenNode,
          mark options={fill=QwenNode, draw=QwenNode},
          mark=*, mark size=1.5
        ] coordinates {
          (8.03, 58.18) (15.17, 49.09) (15.59, 56.36) (11.31, 54.55) 
          (10.51, 60.0) (12.91, 56.36) (16.72, 47.27) (12.47, 60.0) 
          (12.06, 50.91) (15.4, 54.55)
          (24.6, 45.9)
        };
        
        \addplot[only marks, AFNode, mark options={fill=AFNode, draw=AFNode}, mark=star, mark size=2.5] coordinates {(24.4, 45.8)};
        \addplot[only marks, PromptNode, mark options={fill=PromptNode, draw=PromptNode}, mark=diamond*, mark size=2.5] coordinates {(10.7, 45.5)};
        
        \addplot[imitationLine, mark options={fill=imitationNode, draw=imitationNode},  mark=*, mark size=0, dashed, thick] coordinates {(10.23, 65.45) (12.47, 60.0) (15.59, 56.36) (24.6, 45.9)};

        \end{groupplot}

        \node [anchor=south] at ([yshift=0cm] $(group c1r1.north west)!0.5!(group c3r1.north east)$ ) {\pgfplotslegendfromname{sharedlegend}};

        \node [anchor=north] at ([yshift=-0.32cm] group c1r1.south) {\scriptsize Privacy Score (\%)};
        \node [anchor=north] at ([yshift=-0.32cm] group c2r1.south) {\scriptsize Privacy Score (\%)};
        \node [anchor=north] at ([yshift=-0.28cm] group c3r1.south) {\scriptsize Privacy Score (\%)};

        \node [anchor=north] at ([yshift=-0.65cm] $(group c1r1.south)$ ) {\small \textbf{(a)} TAB};
        \node [anchor=north] at ([yshift=-0.65cm] $(group c2r1.south)$ ) {\small \textbf{(b)} SynthPAI};
        \node [anchor=north] at ([yshift=-0.65cm] $(group c3r1.south)$ ) {\small \textbf{(c)} MedQA};

    \end{tikzpicture}
    \vspace{-2em}
    \caption{Trade-off frontier visualization on three datasets across optimized models. Each point represents a distinct anonymization prompt. The \dashuline{dashed line} connects the overall Pareto-optimal solutions across all models, demonstrating the framework's ability to discover diverse privacy-utility trade-offs in a single optimization run.}
    \vspace{-1.1em}
    \label{fig:trade_off_discovery}
\end{figure*}
A key advantage of our framework is its ability to discover \textit{multiple} effective anonymization strategies within a single optimization run per model, rather than converging to a single fixed privacy–utility compromise. This property directly addresses a core limitation of prior anonymization systems, which typically expose only one operating point and require costly re-optimization to target alternative privacy budgets.

Unlike fine-tuning–based approaches, where each trade-off point must be stored as a separate model checkpoint, our framework reduces anonymization strategy discovery to a \textit{string-level search}. Each solution along the trade-off frontier is represented as a natural language instruction, making the full candidate set inexpensive to store, inspect, and deploy. 
Figure~\ref{fig:trade_off_discovery} visualizes the discovered trade-off frontiers on \textsc{TAB}, \textsc{SynthPAI}, and \textsc{MedQA}. Each point corresponds to a distinct anonymization prompt, and the dashed line denotes the global frontier across all models. Across datasets, our framework consistently uncovers a spectrum of solutions spanning privacy-focused configurations (high privacy, reduced utility) and utility-preserving configurations (high utility, moderate privacy), all obtained within a single optimization run.

The discovered frontiers further reveal model-specific trade-off characteristics. On \textsc{TAB}, Qwen3-30B dominates the high-utility region of the frontier, while Mistral-small and Gemma-3-27B occupy competitive regions at lower utility levels. On \textsc{SynthPAI} and \textsc{MedQA}, where privacy and utility are more tightly coupled, the frontier is steeper, yet the optimizer still identifies distinct operating points that would be inaccessible under a single-prompt paradigm. 
From a deployment perspective, practitioners are no longer constrained to choose a “one-size-fits-all” strategy; instead, they can select from a \textit{menu of validated anonymization prompts} depending on regulatory requirements, risk tolerance, or downstream task priorities.

\subsection{Qualitative Analysis}
\begin{figure}[!t]
    \centering
    \small
    \fbox{\parbox{0.8\columnwidth}{
    \scriptsize
Generate a new anonymized version that achieves maximum privacy through comprehensive de-identification and stylistic obfuscation, while preserving full diagnostic and clinical reasoning fidelity. $[...]$ Preserve Diagnostic Fidelity: Do not remove, generalize, or obscure any clinically diagnostic fact essential to arriving at the correct answer. Retain key diagnostic landmarks exactly as they inform the diagnosis, \hlblue{including pathognomonic findings, critical lab relationships, and temporal or anatomical patterns.} $[...]$ Maximize Anonymization Through De-identification: Replace all personally identifiable information and quasi-identifiers (age, gender, ethnicity, location) with generalized, non-specific alternatives, avoiding high-risk combinations of descriptors. $[...]$ Disrupt Writing Style Aggressively to Maximize Privacy: The anonymized text must be unrecognizable in style from the original. \hlblue{Apply three or more stylistic transformations, such as sentence restructuring, passive voice, embedded clauses, and lexical substitution,} while never altering diagnostic anchors. $[...]$ Final validation: Would a trained physician still be able to identify the correct diagnosis? Would a writing-style analyzer fail to detect this as a near-clone? \hlblue{If either answer is no, the anonymization has failed.} }}
    \caption{Optimized anonymization prompt for \Qwenemoji{}~Qwen3-30B-A3B on the MedQA task.}
    \label{fig:medqa_prompt}
    \vspace{-1em}
\end{figure}

\begin{table*}[t]
\small
\centering
\begin{tabularx}{\textwidth}{l c c X}
\toprule
\textbf{Model} & \textbf{Priv.} & \textbf{Util.} & \textbf{Anonymized Text} \\
\midrule
Original & --- & --- & hit up charity shops for books \& clothes - you'd be amazed at some finds! also big into meal prepping on sundays saves me tons during busy \textbf{uni days} \\
\midrule
\Qwenemoji{}~Qwen3-30B-A3B & 1.0 \cmark & 0.87 \cmark & hit up secondhand stores for books \& clothes - you'd be amazed at some finds! also big into meal prepping on Sundays saves me tons during busy \textbf{weekdays} \\
\midrule
\Googleemoji{}~Gemma-3-27B-it & 0.0 \xmark & 0.73 \cmark & I like to shop at thrift stores for books \& clothes - you'd be amazed at some finds! Also big into meal prepping on weekends, saves me a lot during busy \textbf{school days} \\
\midrule
\pixtralemoji{}~Mistral-Small-3.2-24B & 0.0 \xmark & 0.52 \xmark & Check out second-hand stores for books and clothing - you might be surprised by some of the treasures! Also, big into preparing meals on Sundays saves me a lot during hectic \textbf{academic days} \\
\bottomrule
\end{tabularx}
\caption{Qualitative comparison on \textsc{SynthPAI} test sample. The original text reveals the author's \textit{education} (ground truth: bachelor's degree).}
\label{tab:qualitative_synthpai_2}
\vspace{-1em}
\end{table*}

Figure~\ref{fig:medqa_prompt} presents a representative optimized anonymization prompt discovered on MedQA using \Qwenemoji{}~Qwen3-30B-A3B. The prompt explicitly separates diagnostic invariance from stylistic obfuscation, enforcing strong privacy protection against LLM-based inference attacks while preserving clinical reasoning. Diagnostic fidelity is maintained by requiring the retention of pathognomonic findings, clinically meaningful relationships, and temporal or anatomical patterns that are essential for arriving at the correct diagnosis.

From a privacy perspective, the prompt goes beyond standard de-identification by aggressively targeting stylometric leakage. It combines removal of explicit identifiers with structural rewrites, syntactic variation, and lexical substitution of non-essential content, yielding anonymized outputs that are difficult to link to the original text through similarity or authorship analysis. Overall, this example illustrates how explicitly encoding both privacy and utility failure modes in the optimization objective enables LLMs to discover anonymization strategies that effectively balance anonymity and clinical fidelity.

Table~\ref{tab:qualitative_synthpai_2} illustrates a localized privacy failure on \textsc{SynthPAI}. The target attribute is \textit{education} (ground truth: bachelor's degree), and the leakage is concentrated in the colloquial expression ``uni days'' whose association with university study is sufficiently strong for the attacker LLM to recover the label. Qwen3 successfully removes this cue by replacing it with the generic term ``weekdays'' eliminating the educational signal while preserving the surrounding content almost verbatim (ROUGE-1 $= 0.87$). In contrast, both alternative models retain educational semantics despite lexical variation. Gemma rewrites the phrase as ``school days'' and Mistral-Small as ``academic days''; although neither is a literal copy, both remain anchored in the education domain and therefore result in privacy failure. Mistral further introduces broader paraphrasing elsewhere in the sentence which substantially lowers utility (ROUGE-1 $= 0.52$) without improving privacy. Overall, this example shows that successful anonymization in \textsc{SynthPAI} does not necessarily require extensive rewriting, but rather precise removal of the minimal lexical cue that reveals the target attribute, reflecting a targeted form of semantic substitution learned by optimized models.

\section{Conclusion}
We introduce \textit{adaptive text anonymization} as a paradigm that leverages large language models to learn task–specific privacy–utility trade-offs. We propose a new framework that formulates anonymization as a string discovery problem, extending GEPA with a three-stage optimization pipeline enabling a better exploration of trade-offs and the discovery of multiple Pareto-optimal solutions within a single optimization run. Experiments across five datasets show that our approach consistently learns effective trade-offs and compares favorably to existing LLM-based anonymization methods. Together, these results highlight the potential of adaptive, model-driven anonymization for robust and flexible privacy preservation across domains.

\section*{Limitations}
While our framework demonstrates strong performance across diverse anonymization tasks, several limitations remain and highlight directions for future work. 

First, privacy and utility are primarily combined using an unweighted metric aggregation. Although this formulation is sufficient to expose a broad range of trade-offs, it does not explicitly model alternative decision rules such as lexicographic ordering, weighted objectives, or hard privacy constraints. Exploring these formulations could further align the optimization process with domain- or regulation-specific requirements.

Second, our approach requires a small annotated training and validation set for each task to guide optimization. This introduces additional computational overhead compared to fully zero-shot anonymization pipelines. We view this requirement as a reasonable trade-off: even limited supervision enables the discovery of substantially stronger privacy–utility operating points that allow smaller, local models to compete with larger closed-source systems. Nonetheless, reducing this supervision requirement remains an important direction for future work.

Third, although a key contribution of our framework is enabling anonymization with locally deployable open-source models, the evaluation pipeline still relies on closed-source LLMs for certain privacy and utility metrics. This reliance is partially mitigated by the small number of annotated examples required during optimization, which limits the amount of sensitive data that must leave the local boundary. However, determining which anonymized text is itself safe to share with external evaluators remains and is not fully resolved in this work.

Fourth, we exclude reasoning-oriented models (e.g., relying on extended chain-of-thought inference at test time) from our study. Although such models could, in principle, derive effective anonymization strategies via in-context reasoning, they would need to re-derive these strategies independently for each input, resulting in substantially higher per-instance computational costs. In addition, current reasoning models typically require very large parameter scales to robustly address complex heterogeneous domains. Exploring whether smaller reasoning models, or hybrid approaches that combine prompt optimization with limited test-time reasoning, can provide complementary advantages is an intriguing direction for future work.

Finally, the inherent non-determinism of LLM generation can introduce instability during optimization, occasionally leading to variance in training trajectories or convergence behavior. While evolutionary optimization is relatively robust to such noise when the rollout budget is large enough, improving stability and reproducibility remains an open challenge.

\section*{Ethical Considerations}
This work addresses a socially important problem: enabling the safe sharing and analysis of sensitive textual data while preserving individual privacy. At the same time, our approach brings forward ethical considerations that should be considered in broader or higher-stakes deployments. First, reliance on automated LLM-based evaluators may obscure certain failure modes or systematically underweight rare but consequential privacy leaks. Second, optimized anonymization prompts could be misused to intentionally obfuscate accountability, attribution, or provenance in contexts where transparency is required. Third, stylometric and contextual obfuscation, while effective for privacy protection, may also remove signals that are valuable for forensic analysis, content moderation, or safety monitoring. Finally, heavy reliance on LLM-based evaluation introduces the risk of encoding fairness evaluator biases and instability.

These risks highlight the importance of treating anonymization as a risk-reduction mechanism rather than a guaranty of privacy. Even after anonymization, residual risks may remain, particularly in domains involving sensitive health, legal, or personal data. Deployment of our framework should therefore be accompanied by explicit consent considerations, clear articulation of intended data use, and alignment with applicable regulatory frameworks such as GDPR or HIPAA.

\bibliography{custom,anthology-1,anthology-2}

\begin{thebibliography}{40}
\providecommand{\natexlab}[1]{#1}

\bibitem[{Agrawal et~al.(2026)Agrawal, Tan, Soylu, Ziems, Khare, Opsahl-Ong, Singhvi, Shandilya, Ryan, Jiang, Potts, Sen, Dimakis, Stoica, Klein, Zaharia, and Khattab}]{agrawal2025gepareflectivepromptevolution}
Lakshya~A Agrawal, Shangyin Tan, Dilara Soylu, Noah Ziems, Rishi Khare, Krista Opsahl-Ong, Arnav Singhvi, Herumb Shandilya, Michael~J Ryan, Meng Jiang, Christopher Potts, Koushik Sen, Alex Dimakis, Ion Stoica, Dan Klein, Matei Zaharia, and Omar Khattab. 2026.
\newblock \href {https://openreview.net/forum?id=RQm2KQTM5r} {{GEPA}: Reflective prompt evolution can outperform reinforcement learning}.
\newblock In \emph{The Fourteenth International Conference on Learning Representations}.

\bibitem[{Brahem et~al.(2024)Brahem, Watissee, Eichler, Boiret, Anciaux, and Maria~de Fuentes}]{brahem:hal-04684512}
Mariem Brahem, Jasmine Watissee, C{\'e}dric Eichler, Adrien Boiret, Nicolas Anciaux, and Jose Maria~de Fuentes. 2024.
\newblock \href {https://inria.hal.science/hal-04684512} {retellme: Design rules for using large language models to protect the privacy of individuals in their textual contributions}.
\newblock In \emph{DPM 2024 - International Workshop on Data Privacy Management @ ESORICS}, Barcelona, Spain.

\bibitem[{Chen et~al.(2024)Chen, Chen, Goldstein, Huang, and Zhou}]{chen2024instructzero}
Lichang Chen, Jiuhai Chen, Tom Goldstein, Heng Huang, and Tianyi Zhou. 2024.
\newblock \href {https://openreview.net/forum?id=rADFNrIss3} {Instructzero: Efficient instruction optimization for black-box large language models}.
\newblock In \emph{Forty-first International Conference on Machine Learning}.

\bibitem[{Comanici et~al.(2025)Comanici, Bieber, Schaekermann, Pasupat, Sachdeva, Dhillon, Blistein, Ram, Zhang, Rosen, Marris, Petulla, Gaffney, Aharoni, Lintz, Pais, Jacobsson, Szpektor, Jiang, Haridasan, Omran, Saunshi, Bahri, Mishra, Chu, Boyd, Hekman, Parisi, Zhang, Kawintiranon, Bedrax-Weiss, Wang, Xu, Purkiss, Mendlovic, Deutel, Nguyen, Langley, Korn, Rossazza, Ramé, Waghmare, Miller, Byrd, Sheshan, Hadsell, Bhardwaj, Janus, Rissa, Horgan, Abdagic, Belenki, Allingham, Singh, Guidroz, Srinivasan, Schmit, Chiafullo, Elisseeff, Jha, Kolhar, Berrada, Ding, Si, Mallick, Och, Erell, Ni, Latkar, Yang, Sirkovic, Feng, Leland, Hornung, Wu, Blundell, Alvari, Huang, Yip, Deur, Liu, Surita, Duque, Damen, Jia, Guez, Mircea, Sinha, Magni, Stradomski, Marian, Galić, Chen, Husain, Singhal, Grewe, Aubet, Song, Blanco, Rechis, Ho, Munoz, Zheng, Hamrick, Mather, Taitelbaum, Rutherford, Lei, Chen, Shukla, Moreira, Doi, Isik, Shabat, Rogozińska, Kolipaka, Chang, Vušak, Venkatachary, Noghabi, Bharti, Jun, Zaks, Green,
  Challagundla, Wong, Mohammad, Hirsch, Cheng, Naim, Proleev, Vincent, Singh, Krikun, Krishnan, Ghahramani, Atias, Aggarwal, Kirov, Vytiniotis, Koh, Chronopoulou, Dogra, Ion, Tyen, Lee, Weissenberger, Strohman, Balakrishna, Rae, Velic, de~Liedekerke, Elyada, Yuan, Liu, Shani, Kishchenko, Alessio, Li, Song, Kwei, Jankowski, Pappu, Namiki, Ma, Tripuraneni, Cherry, Ikonomidis, Ling, Ji, Westberg, Wright, Yu, Parkinson, Ramaswamy, Connor, Yeganeh, Grover, Kenwright, Litchev, Apps, Tomala, Halim, Castro-Ros, Li, Boral, Sho, Yarom, Malmi, Klinghoffer, Lin, Ansell, S, Zhao, Zuo, Santoro, Cheng, Demmessie, Liu, Brichtova, Culp, Braun, Graur, Ng, Mehta, Phillips, Sundberg, Godbole, Liu, Katariya, Rim, Seyedhosseini, Ammirati, Valfridsson, Malihi, Knight, Toor, Lampe, Ittycheriah, and et~al.}]{comanici2025gemini25pushingfrontier}
Gheorghe Comanici, Eric Bieber, Mike Schaekermann, Ice Pasupat, Noveen Sachdeva, Inderjit Dhillon, Marcel Blistein, Ori Ram, Dan Zhang, Evan Rosen, Luke Marris, Sam Petulla, Colin Gaffney, Asaf Aharoni, Nathan Lintz, Tiago~Cardal Pais, Henrik Jacobsson, Idan Szpektor, Nan-Jiang Jiang, and 201 others. 2025.
\newblock \href {https://arxiv.org/abs/2507.06261} {Gemini 2.5: Pushing the frontier with advanced reasoning, multimodality, long context, and next generation agentic capabilities}.
\newblock \emph{Preprint}, arXiv:2507.06261.

\bibitem[{Deußer et~al.(2025)Deußer, Sparrenberg, Berger, Hahnbück, Bauckhage, and Sifa}]{deusser2025survey}
Tobias Deußer, Lorenz Sparrenberg, Armin Berger, Max Hahnbück, Christian Bauckhage, and Rafet Sifa. 2025.
\newblock \href {https://doi.org/10.1109/DSAA65442.2025.11247969} {A survey on current trends and recent advances in text anonymization}.
\newblock In \emph{2025 IEEE 12th International Conference on Data Science and Advanced Analytics (DSAA)}, pages 1--9.

\bibitem[{Dou et~al.(2024)Dou, Krsek, Naous, Kabra, Das, Ritter, and Xu}]{dou-etal-2024-reducing}
Yao Dou, Isadora Krsek, Tarek Naous, Anubha Kabra, Sauvik Das, Alan Ritter, and Wei Xu. 2024.
\newblock \href {https://doi.org/10.18653/v1/2024.acl-long.741} {Reducing privacy risks in online self-disclosures with language models}.
\newblock In \emph{Proceedings of the 62nd Annual Meeting of the Association for Computational Linguistics (Volume 1: Long Papers)}, pages 13732--13754, Bangkok, Thailand. Association for Computational Linguistics.

\bibitem[{Dwork(2006)}]{dwork2006dp}
Cynthia Dwork. 2006.
\newblock Differential privacy.
\newblock In \emph{Automata, Languages and Programming}, pages 1--12, Berlin, Heidelberg. Springer Berlin Heidelberg.

\bibitem[{Francopoulo and Schaub(2020)}]{francopoulo2020anonymization}
Gil Francopoulo and L{\'e}on-Paul Schaub. 2020.
\newblock \href {https://hal.science/hal-02939437} {{Anonymization for the GDPR in the Context of Citizen and Customer Relationship Management and NLP}}.
\newblock In \emph{{workshop on Legal and Ethical Issues (Legal2020)}}, pages 9--14, Marseille, France. {LREC2020}, {ELRA}.

\bibitem[{Frikha et~al.(2024)Frikha, Walha, Nakka, Mendes, Jiang, and Zhou}]{frikha2024incognitext}
Ahmed Frikha, Nassim Walha, Krishna~Kanth Nakka, Ricardo Mendes, Xue Jiang, and Xuebing Zhou. 2024.
\newblock \href {https://openreview.net/forum?id=JRifjkHove} {Incognitext: Privacy-enhancing conditional text anonymization via {LLM}-based private attribute randomization}.
\newblock In \emph{Neurips Safe Generative AI Workshop 2024}.

\bibitem[{{GDPR}(2016)}]{GDPRArticle5}
{GDPR}. 2016.
\newblock \href {https://gdpr-info.eu/art-5-gdpr/} {Article 5 — principles relating to processing of personal data}.
\newblock Accessed: 2025-12-16.

\bibitem[{Hathurusinghe et~al.(2021)Hathurusinghe, Nejadgholi, and Bolic}]{hathurusinghe-etal-2021-privacy}
Rajitha Hathurusinghe, Isar Nejadgholi, and Miodrag Bolic. 2021.
\newblock \href {https://doi.org/10.18653/v1/2021.privatenlp-1.5} {A privacy-preserving approach to extraction of personal information through automatic annotation and federated learning}.
\newblock In \emph{Proceedings of the Third Workshop on Privacy in Natural Language Processing}, pages 36--45, Online. Association for Computational Linguistics.

\bibitem[{Jin et~al.(2021)Jin, Pan, Oufattole, Weng, Fang, and Szolovits}]{jin2021disease}
Di~Jin, Eileen Pan, Nassim Oufattole, Wei-Hung Weng, Hanyi Fang, and Peter Szolovits. 2021.
\newblock What disease does this patient have? a large-scale open domain question answering dataset from medical exams.
\newblock \emph{Applied Sciences}, 11(14):6421.

\bibitem[{Khattab et~al.(2024)Khattab, Singhvi, Maheshwari, Zhang, Santhanam, A, Haq, Sharma, Joshi, Moazam, Miller, Zaharia, and Potts}]{khattab2024dspy}
Omar Khattab, Arnav Singhvi, Paridhi Maheshwari, Zhiyuan Zhang, Keshav Santhanam, Sri~Vardhamanan A, Saiful Haq, Ashutosh Sharma, Thomas~T. Joshi, Hanna Moazam, Heather Miller, Matei Zaharia, and Christopher Potts. 2024.
\newblock \href {https://openreview.net/forum?id=sY5N0zY5Od} {{DSP}y: Compiling declarative language model calls into state-of-the-art pipelines}.
\newblock In \emph{The Twelfth International Conference on Learning Representations}.

\bibitem[{Li et~al.(2025)Li, Raghuram, Khattab, Hirschberg, and Yu}]{li-etal-2025-papillon}
Siyan Li, Vethavikashini~Chithrra Raghuram, Omar Khattab, Julia Hirschberg, and Zhou Yu. 2025.
\newblock \href {https://doi.org/10.18653/v1/2025.naacl-long.173} {{PAPILLON}: Privacy preservation from {I}nternet-based and local language model ensembles}.
\newblock In \emph{Proceedings of the 2025 Conference of the Nations of the Americas Chapter of the Association for Computational Linguistics: Human Language Technologies (Volume 1: Long Papers)}, pages 3371--3390, Albuquerque, New Mexico. Association for Computational Linguistics.

\bibitem[{Lison et~al.(2021)Lison, Pil{\'a}n, Sanchez, Batet, and {\O}vrelid}]{lison-etal-2021-anonymisation}
Pierre Lison, Ildik{\'o} Pil{\'a}n, David Sanchez, Montserrat Batet, and Lilja {\O}vrelid. 2021.
\newblock \href {https://doi.org/10.18653/v1/2021.acl-long.323} {Anonymisation models for text data: State of the art, challenges and future directions}.
\newblock In \emph{Proceedings of the 59th Annual Meeting of the Association for Computational Linguistics and the 11th International Joint Conference on Natural Language Processing (Volume 1: Long Papers)}, pages 4188--4203, Online. Association for Computational Linguistics.

\bibitem[{Loiseau et~al.(2025)Loiseau, Sileo, Riquet, Meyer, and Tommasi}]{loiseau-etal-2025-tau}
Gabriel Loiseau, Damien Sileo, Damien Riquet, Maxime Meyer, and Marc Tommasi. 2025.
\newblock \href {https://doi.org/10.18653/v1/2025.emnlp-demos.16} {Tau-eval: A unified evaluation framework for useful and private text anonymization}.
\newblock In \emph{Proceedings of the 2025 Conference on Empirical Methods in Natural Language Processing: System Demonstrations}, pages 216--227, Suzhou, China. Association for Computational Linguistics.

\bibitem[{Mattern et~al.(2022)Mattern, Weggenmann, and Kerschbaum}]{mattern-etal-2022-limits}
Justus Mattern, Benjamin Weggenmann, and Florian Kerschbaum. 2022.
\newblock \href {https://doi.org/10.18653/v1/2022.findings-naacl.65} {The limits of word level differential privacy}.
\newblock In \emph{Findings of the Association for Computational Linguistics: NAACL 2022}, pages 867--881, Seattle, United States. Association for Computational Linguistics.

\bibitem[{Meisenbacher et~al.(2025)Meisenbacher, Chevli, and Matthes}]{meisenbacher-etal-2025-impact}
Stephen Meisenbacher, Maulik Chevli, and Florian Matthes. 2025.
\newblock \href {https://doi.org/10.18653/v1/2025.findings-naacl.32} {On the impact of noise in differentially private text rewriting}.
\newblock In \emph{Findings of the Association for Computational Linguistics: NAACL 2025}, pages 514--532, Albuquerque, New Mexico. Association for Computational Linguistics.

\bibitem[{Meisenbacher et~al.(2024)Meisenbacher, Chevli, Vladika, and Matthes}]{meisenbacher-etal-2024-dp}
Stephen Meisenbacher, Maulik Chevli, Juraj Vladika, and Florian Matthes. 2024.
\newblock \href {https://doi.org/10.18653/v1/2024.findings-acl.554} {{DP}-{MLM}: Differentially private text rewriting using masked language models}.
\newblock In \emph{Findings of the Association for Computational Linguistics: ACL 2024}, pages 9314--9328, Bangkok, Thailand. Association for Computational Linguistics.

\bibitem[{{OpenAI}(2025)}]{gpt5systemcard}
{OpenAI}. 2025.
\newblock \href {https://cdn.openai.com/gpt-5-system-card.pdf} {Gpt-5 system card}.
\newblock Technical report, OpenAI.
\newblock Accessed: 2025-12-18.

\bibitem[{Opsahl-Ong et~al.(2024)Opsahl-Ong, Ryan, Purtell, Broman, Potts, Zaharia, and Khattab}]{opsahl-ong-etal-2024-optimizing}
Krista Opsahl-Ong, Michael~J Ryan, Josh Purtell, David Broman, Christopher Potts, Matei Zaharia, and Omar Khattab. 2024.
\newblock \href {https://doi.org/10.18653/v1/2024.emnlp-main.525} {Optimizing instructions and demonstrations for multi-stage language model programs}.
\newblock In \emph{Proceedings of the 2024 Conference on Empirical Methods in Natural Language Processing}, pages 9340--9366, Miami, Florida, USA. Association for Computational Linguistics.

\bibitem[{Pasch and Cha(2025)}]{pasch-cha-2025-balancing}
Stefan Pasch and Min~Chul Cha. 2025.
\newblock \href {https://doi.org/10.18653/v1/2025.privatenlp-main.3} {Balancing privacy and utility in personal {LLM} writing tasks: An automated pipeline for evaluating anonymizations}.
\newblock In \emph{Proceedings of the Sixth Workshop on Privacy in Natural Language Processing}, pages 32--41, Albuquerque, New Mexico. Association for Computational Linguistics.

\bibitem[{Patsakis and Lykousas(2023)}]{patsakis2023man}
Constantinos Patsakis and Nikolaos Lykousas. 2023.
\newblock Man vs the machine in the struggle for effective text anonymisation in the age of large language models.
\newblock In \emph{Scientific Reports}, volume~13, page 16026. Nature Publishing Group UK London.

\bibitem[{Pil{\'a}n et~al.(2022)Pil{\'a}n, Lison, {\O}vrelid, Papadopoulou, S{\'a}nchez, and Batet}]{pilan-etal-2022-text}
Ildik{\'o} Pil{\'a}n, Pierre Lison, Lilja {\O}vrelid, Anthi Papadopoulou, David S{\'a}nchez, and Montserrat Batet. 2022.
\newblock \href {https://doi.org/10.1162/coli_a_00458} {The text anonymization benchmark ({TAB}): A dedicated corpus and evaluation framework for text anonymization}.
\newblock \emph{Computational Linguistics}, 48(4):1053--1101.

\bibitem[{Rivera-Soto et~al.(2021)Rivera-Soto, Miano, Ordonez, Chen, Khan, Bishop, and Andrews}]{rivera-soto-etal-2021-learning}
Rafael~A. Rivera-Soto, Olivia~Elizabeth Miano, Juanita Ordonez, Barry~Y. Chen, Aleem Khan, Marcus Bishop, and Nicholas Andrews. 2021.
\newblock \href {https://doi.org/10.18653/v1/2021.emnlp-main.70} {Learning universal authorship representations}.
\newblock In \emph{Proceedings of the 2021 Conference on Empirical Methods in Natural Language Processing}, pages 913--919, Online and Punta Cana, Dominican Republic. Association for Computational Linguistics.

\bibitem[{Staab et~al.(2024)Staab, Vero, Balunovic, and Vechev}]{staab2024beyond}
Robin Staab, Mark Vero, Mislav Balunovic, and Martin Vechev. 2024.
\newblock \href {https://openreview.net/forum?id=kmn0BhQk7p} {Beyond memorization: Violating privacy via inference with large language models}.
\newblock In \emph{The Twelfth International Conference on Learning Representations}.

\bibitem[{Staab et~al.(2025)Staab, Vero, Balunovic, and Vechev}]{staab2025anonymizers}
Robin Staab, Mark Vero, Mislav Balunovic, and Martin Vechev. 2025.
\newblock \href {https://openreview.net/forum?id=82p8VHRsaK} {Language models are advanced anonymizers}.
\newblock In \emph{The Thirteenth International Conference on Learning Representations}.

\bibitem[{Team et~al.(2025)Team, Kamath, Ferret, Pathak, Vieillard, Merhej, Perrin, Matejovicova, Ramé, Rivière, Rouillard, Mesnard, Cideron, bastien Grill, Ramos, Yvinec, Casbon, Pot, Penchev, Liu, Visin, Kenealy, Beyer, Zhai, Tsitsulin, Busa-Fekete, Feng, Sachdeva, Coleman, Gao, Mustafa, Barr, Parisotto, Tian, Eyal, Cherry, Peter, Sinopalnikov, Bhupatiraju, Agarwal, Kazemi, Malkin, Kumar, Vilar, Brusilovsky, Luo, Steiner, Friesen, Sharma, Sharma, Gilady, Goedeckemeyer, Saade, Feng, Kolesnikov, Bendebury, Abdagic, Vadi, György, Pinto, Das, Bapna, Miech, Yang, Paterson, Shenoy, Chakrabarti, Piot, Wu, Shahriari, Petrini, Chen, Lan, Choquette-Choo, Carey, Brick, Deutsch, Eisenbud, Cattle, Cheng, Paparas, Sreepathihalli, Reid, Tran, Zelle, Noland, Huizenga, Kharitonov, Liu, Amirkhanyan, Cameron, Hashemi, Klimczak-Plucińska, Singh, Mehta, Lehri, Hazimeh, Ballantyne, Szpektor, Nardini, Pouget-Abadie, Chan, Stanton, Wieting, Lai, Orbay, Fernandez, Newlan, yeong Ji, Singh, Black, Yu, Hui, Vodrahalli, Greff, Qiu,
  Valentine, Coelho, Ritter, Hoffman, Watson, Chaturvedi, Moynihan, Ma, Babar, Noy, Byrd, Roy, Momchev, Chauhan, Sachdeva, Bunyan, Botarda, Caron, Rubenstein, Culliton, Schmid, Sessa, Xu, Stanczyk, Tafti, Shivanna, Wu, Pan, Rokni, Willoughby, Vallu, Mullins, Jerome, Smoot, Girgin, Iqbal, Reddy, Sheth, Põder, Bhatnagar, Panyam, Eiger, Zhang, Liu, Yacovone, Liechty, Kalra, Evci, Misra, Roseberry, Feinberg, Kolesnikov, Han, Kwon, Chen, Chow, Zhu, Wei, Egyed, Cotruta, Giang, Kirk, Rao, Black, Babar, Lo, Moreira, Martins, Sanseviero, Gonzalez, Gleicher, Warkentin, Mirrokni, Senter, Collins, Barral, Ghahramani, Hadsell, Matias, Sculley, Petrov, Fiedel, Shazeer, Vinyals, Dean, Hassabis, Kavukcuoglu, Farabet, Buchatskaya, Alayrac, Anil, Dmitry, Lepikhin, Borgeaud, Bachem, Joulin, Andreev, Hardin, Dadashi, and Hussenot}]{gemmateam2025gemma3technicalreport}
Gemma Team, Aishwarya Kamath, Johan Ferret, Shreya Pathak, Nino Vieillard, Ramona Merhej, Sarah Perrin, Tatiana Matejovicova, Alexandre Ramé, Morgane Rivière, Louis Rouillard, Thomas Mesnard, Geoffrey Cideron, Jean bastien Grill, Sabela Ramos, Edouard Yvinec, Michelle Casbon, Etienne Pot, Ivo Penchev, and 197 others. 2025.
\newblock \href {https://arxiv.org/abs/2503.19786} {Gemma 3 technical report}.
\newblock \emph{Preprint}, arXiv:2503.19786.

\bibitem[{Team(2025)}]{qwen3technicalreport}
Qwen Team. 2025.
\newblock \href {https://arxiv.org/abs/2505.09388} {Qwen3 technical report}.
\newblock \emph{Preprint}, arXiv:2505.09388.

\bibitem[{Utpala et~al.(2023)Utpala, Hooker, and Chen}]{utpala-etal-2023-locally}
Saiteja Utpala, Sara Hooker, and Pin-Yu Chen. 2023.
\newblock \href {https://doi.org/10.18653/v1/2023.findings-emnlp.566} {Locally differentially private document generation using zero shot prompting}.
\newblock In \emph{Findings of the Association for Computational Linguistics: EMNLP 2023}, pages 8442--8457, Singapore. Association for Computational Linguistics.

\bibitem[{Warner et~al.(2025)Warner, Chaffin, Clavi{\'e}, Weller, Hallstr{\"o}m, Taghadouini, Gallagher, Biswas, Ladhak, Aarsen, Adams, Howard, and Poli}]{warner-etal-2025-smarter}
Benjamin Warner, Antoine Chaffin, Benjamin Clavi{\'e}, Orion Weller, Oskar Hallstr{\"o}m, Said Taghadouini, Alexis Gallagher, Raja Biswas, Faisal Ladhak, Tom Aarsen, Griffin~Thomas Adams, Jeremy Howard, and Iacopo Poli. 2025.
\newblock \href {https://doi.org/10.18653/v1/2025.acl-long.127} {Smarter, better, faster, longer: A modern bidirectional encoder for fast, memory efficient, and long context finetuning and inference}.
\newblock In \emph{Proceedings of the 63rd Annual Meeting of the Association for Computational Linguistics (Volume 1: Long Papers)}, pages 2526--2547, Vienna, Austria. Association for Computational Linguistics.

\bibitem[{Wei et~al.(2022)Wei, Wang, Schuurmans, Bosma, Ichter, Xia, Chi, Le, and Zhou}]{10.5555/3600270.3602070}
Jason Wei, Xuezhi Wang, Dale Schuurmans, Maarten Bosma, Brian Ichter, Fei Xia, Ed~H. Chi, Quoc~V. Le, and Denny Zhou. 2022.
\newblock Chain-of-thought prompting elicits reasoning in large language models.
\newblock In \emph{Proceedings of the 36th International Conference on Neural Information Processing Systems}, NIPS '22, Red Hook, NY, USA. Curran Associates Inc.

\bibitem[{Xin et~al.(2025)Xin, Mireshghallah, Li, Duan, Kim, Choi, Tsvetkov, Oh, and Koh}]{xin2025a}
Rui Xin, Niloofar Mireshghallah, Shuyue~Stella Li, Michael Duan, Hyunwoo Kim, Yejin Choi, Yulia Tsvetkov, Sewoong Oh, and Pang~Wei Koh. 2025.
\newblock \href {https://openreview.net/forum?id=oyrDjyasDV} {A false sense of privacy: Evaluating textual data sanitization beyond surface-level privacy leakage}.
\newblock In \emph{ICLR 2025 Workshop on Building Trust in Language Models and Applications}.

\bibitem[{Yang et~al.(2024)Yang, Wang, Lu, Liu, Le, Zhou, and Chen}]{yang2024large}
Chengrun Yang, Xuezhi Wang, Yifeng Lu, Hanxiao Liu, Quoc~V Le, Denny Zhou, and Xinyun Chen. 2024.
\newblock \href {https://openreview.net/forum?id=Bb4VGOWELI} {Large language models as optimizers}.
\newblock In \emph{The Twelfth International Conference on Learning Representations}.

\bibitem[{Yang et~al.(2025)Yang, Zhu, and Gurevych}]{yang-etal-2025-robust}
Tianyu Yang, Xiaodan Zhu, and Iryna Gurevych. 2025.
\newblock \href {https://doi.org/10.18653/v1/2025.acl-long.1404} {Robust utility-preserving text anonymization based on large language models}.
\newblock In \emph{Proceedings of the 63rd Annual Meeting of the Association for Computational Linguistics (Volume 1: Long Papers)}, pages 28922--28941, Vienna, Austria. Association for Computational Linguistics.

\bibitem[{Yermilov et~al.(2023)Yermilov, Raheja, and Chernodub}]{yermilov-etal-2023-privacy}
Oleksandr Yermilov, Vipul Raheja, and Artem Chernodub. 2023.
\newblock \href {https://doi.org/10.18653/v1/2023.trustnlp-1.20} {Privacy- and utility-preserving {NLP} with anonymized data: A case study of pseudonymization}.
\newblock In \emph{Proceedings of the 3rd Workshop on Trustworthy Natural Language Processing (TrustNLP 2023)}, pages 232--241, Toronto, Canada. Association for Computational Linguistics.

\bibitem[{Yukhymenko et~al.(2024)Yukhymenko, Staab, Vero, and Vechev}]{yukhymenko2024synthetic}
Hanna Yukhymenko, Robin Staab, Mark Vero, and Martin Vechev. 2024.
\newblock \href {https://openreview.net/forum?id=1nqfIQIQBf} {A synthetic dataset for personal attribute inference}.
\newblock In \emph{Thirty-eighth Conference on Neural Information Processing Systems Datasets and Benchmarks Track}.

\bibitem[{Zhao et~al.(2024)Zhao, Ren, Hessel, Cardie, Choi, and Deng}]{zhao2024wildchat}
Wenting Zhao, Xiang Ren, Jack Hessel, Claire Cardie, Yejin Choi, and Yuntian Deng. 2024.
\newblock \href {https://openreview.net/forum?id=Bl8u7ZRlbM} {Wildchat: 1m chat{GPT} interaction logs in the wild}.
\newblock In \emph{The Twelfth International Conference on Learning Representations}.

\bibitem[{Zhou et~al.(2023)Zhou, Muresanu, Han, Paster, Pitis, Chan, and Ba}]{zhou2023large}
Yongchao Zhou, Andrei~Ioan Muresanu, Ziwen Han, Keiran Paster, Silviu Pitis, Harris Chan, and Jimmy Ba. 2023.
\newblock \href {https://openreview.net/forum?id=92gvk82DE-} {Large language models are human-level prompt engineers}.
\newblock In \emph{The Eleventh International Conference on Learning Representations}.

\bibitem[{Çano and Habernal(2025)}]{çano2025differentiallyprivatetextgenerationdegrades}
Erion Çano and Ivan Habernal. 2025.
\newblock \href {https://arxiv.org/abs/2509.11176} {Differentially-private text generation degrades output language quality}.
\newblock \emph{Preprint}, arXiv:2509.11176.

\end{thebibliography}

\newpage
\appendix
\section{Prompt Optimization Ablation Study}
\label{apx:ablation_study}
\begin{figure}[t]
    \centering
    \includegraphics[width=\columnwidth]{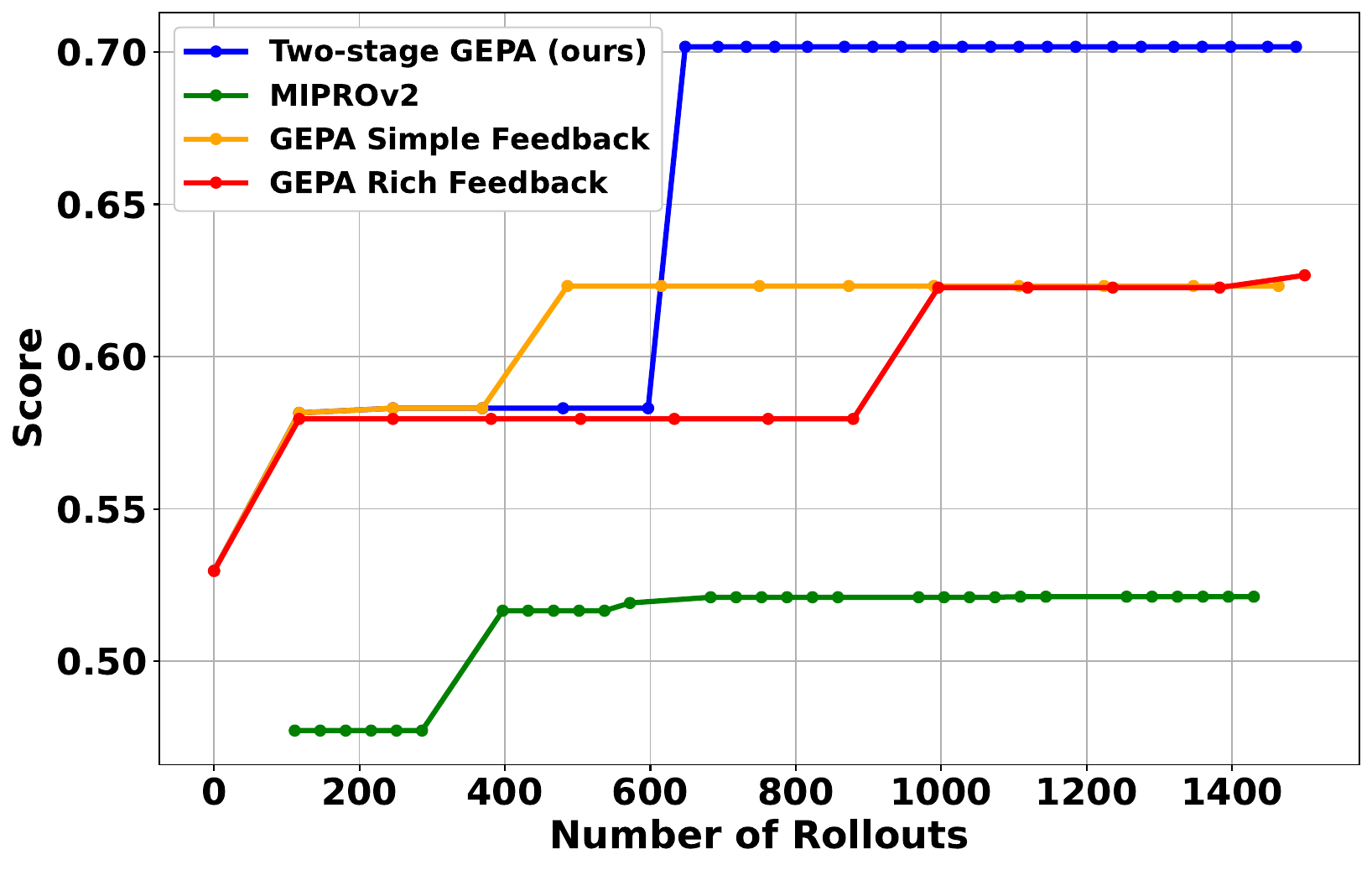}
    \includegraphics[width=\columnwidth]{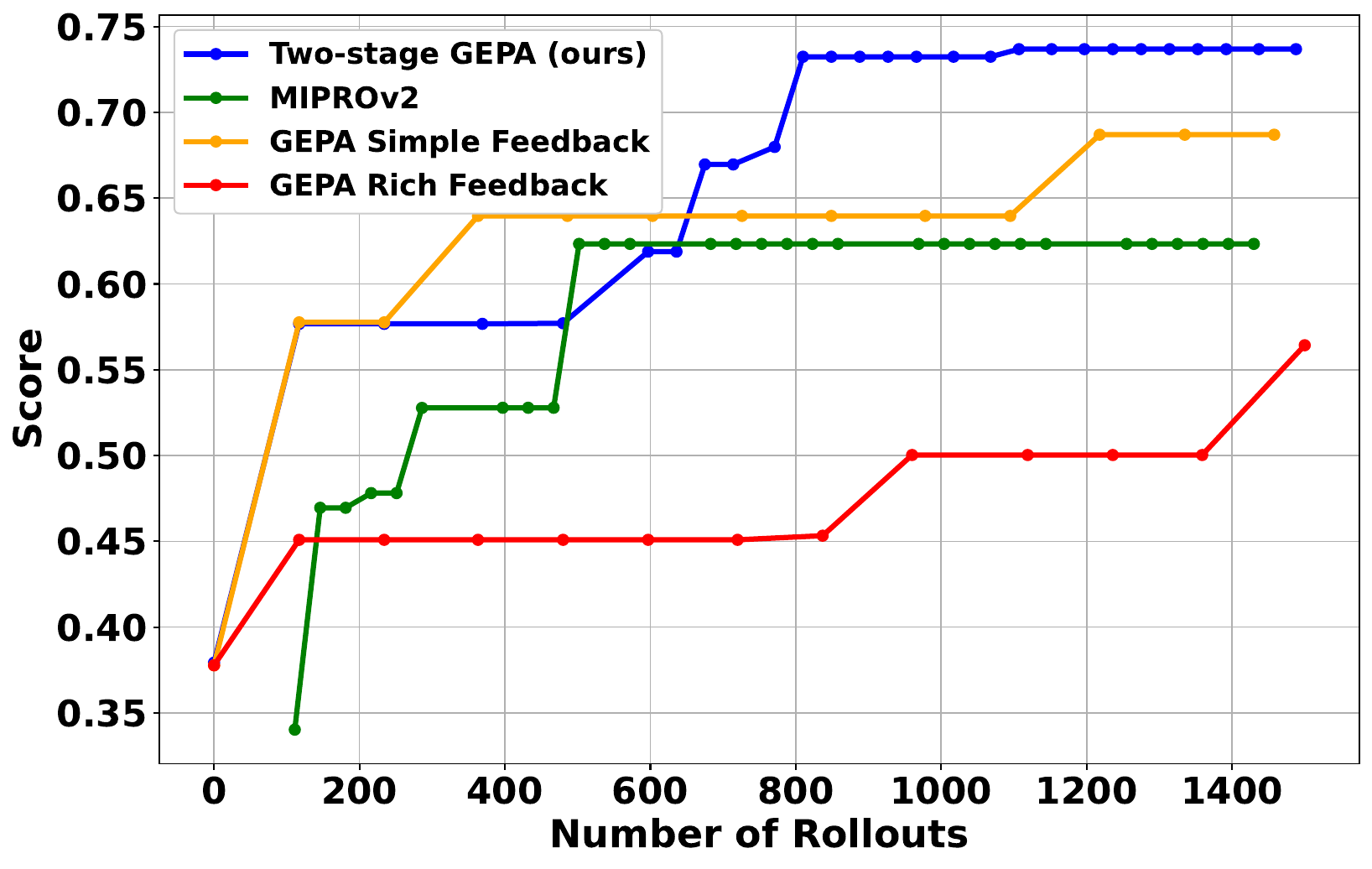}
    \caption{A comparison of learning behavior of our modified GEPA implementation against each separated component and a state-of-the-art prompt optimizer reference (MIPROv2). Results are measures with \Googleemoji{}~Gemma-3-27B-it on SynthPAI (top) and \pixtralemoji{}~Mistral-Small-3.2-24B on TAB (bottom).}
    \label{fig:ablation_learning_curves}
\end{figure}

Figure~\ref{fig:ablation_learning_curves} ablates the main components of our GEPA-based optimizer by comparing: (i) our full \textbf{two-stage GEPA} pipeline (warm-start + refinement), (ii) a \textbf{warm-start-only} variant using \textit{simple} scalar feedback (\textsc{GEPA Simple Feedback}), (iii) a \textbf{refinement-only} variant driven by \textit{rich} structured feedback (\textsc{GEPA Rich Feedback}), and (iv) the state-of-the-art prompt optimizer \textsc{MIPROv2}~\cite{opsahl-ong-etal-2024-optimizing}. All methods are run under the same rollout budget and evaluated using the same task score as in the main paper. We report learning curves on two representative tasks/backbones: \textsc{SynthPAI} with Gemma-3-27B-it (top) and \textsc{TAB} with Mistral-Small-3.2-24B (bottom).

Across both tasks, the full two-stage method is consistently the best final performer and the most reliable. On \textsc{SynthPAI}, \textsc{GEPA Simple Feedback} makes rapid early progress but quickly saturates (plateauing around $\sim$0.62), while \textsc{GEPA Rich Feedback} improves more gradually and reaches a similar plateau. In contrast, \textbf{two-stage GEPA} combines the fast initial gains of warm-start with a decisive refinement jump, ultimately reaching $\sim$0.70 and maintaining that advantage over the remainder of the budget. This pattern suggests the warm-start stage efficiently finds a strong region of the prompt space, while refinement is necessary to escape the early plateau and reach higher-quality trade-offs.

On \textsc{TAB}, the separation between components is even clearer. \textsc{GEPA Simple Feedback} again improves quickly (reaching $\sim$0.63 early and later $\sim$0.69), whereas \textsc{GEPA Rich Feedback} alone lags substantially and only recovers late in the budget (ending around $\sim$0.56). Our method steadily accumulates improvements throughout optimization and achieves the best final score (about $\sim$0.74), outperforming both single-component variants and \textsc{MIPROv2} (which plateaus around $\sim$0.62). Overall, these results highlight that warm-start is important for sample-efficient early gains, refinement provides the additional progress needed to surpass strong plateaus, and combining the two stages yields a robust optimizer that dominates a competitive baseline under a relatively small and fixed compute budget.

\section{Backbone Evaluation Robustness}
On our paper, all evaluations that require an LLM-based judge are conducted using \Googleemoji{}~Gemini-2.5-flash. To contextualize this choice and sanity-check that our reported improvements are not an artifact of unusually permissive judging, we run the same evaluation pipeline on the original, non-anonymized inputs. Table~\ref{tab:gemini_results} reports these baselines \texttt{privacy/utility} scores.

As expected, the unmodified texts exhibit near-maximal privacy leakage across tasks: privacy is close to zero for \textsc{SynthPAI}, \textsc{TAB}, and \textsc{MedQA}, indicating that the corresponding privacy objectives are achieved almost systematically without any protective rewriting. \textsc{DB-Bio} and \textsc{PUPA} show slightly higher privacy values, but remain very low, reflecting that these datasets still contain substantial identifying content even before any adversarial probing. In contrast, utility remains high on tasks where the metric measures task performance or content fidelity without requiring anonymization (e.g. \textsc{DB-Bio} and \textsc{PUPA}), confirming that the raw texts are well-formed for downstream use and that the evaluation is capable of recognizing strong task signal.
\begin{table}[t]
    \centering
    \setlength{\tabcolsep}{3pt} 
    \renewcommand{\arraystretch}{1.2}
    
    \resizebox{\columnwidth}{!}{
    \begin{tabular}{lccccc}
    \textbf{Methods}
     & \textbf{DB-Bio} 
     & \textbf{SynthPAI}
     & \textbf{TAB}
     & \textbf{PUPA} 
     & \textbf{MedQA} \\
    \midrule
    \textit{No Anonymization} & 6.29$^\dagger$/100$^\dagger$ & 0$^\dagger$/100 & 0/100 & 5.62$^\dagger$/99.2$^\dagger$ & 0/64.0$^\dagger$ \\
    \bottomrule
    \end{tabular}
    }
    \caption{Evaluation results on original texts (encoded as \texttt{privacy/utility}). Tasks requiring a backbone LLM (\Googleemoji{}~Gemini-2.5-flash) are marked with $\dagger$.}
    \label{tab:gemini_results}
\end{table}

\section{Model Scaling}
\label{apx:slm_study}
We study whether the benefits of adaptive text anonymization persist on smaller language models by repeating the same prompt-optimization procedure on a substantially smaller one (\Qwenemoji{}~Qwen-2.5-7B) compared to the mid-size model \Qwenemoji{}~Qwen3-30B-A3B used in our core paper. Table~\ref{tab:slm_results} reports \texttt{privacy/utility} for the \textit{Seed Prompt} and the corresponding \textit{Optimized Prompt} on all five tasks.

Overall, optimization yields consistent privacy gains at both scales, indicating that instruction learning transfers beyond large backbones. For \Qwenemoji{}~Qwen-2.5-7B: optimization delivers substantial privacy improvements on \textsc{SynthPAI} (5.88$\rightarrow$17.6) and \textsc{TAB} (34.3$\rightarrow$90.6), and moderate gains on \textsc{DB-Bio} and \textsc{PUPA}. While the smaller model typically starts from a weaker seed prompt (lower privacy and sometimes lower utility than the 30B model), the optimized prompts recover a fraction of the privacy gains observed at 30B: most strikingly on \textsc{TAB}. That said, the 7B model more often exhibits stronger privacy--utility coupling (e.g., \textsc{SynthPAI} and \textsc{PUPA}), suggesting that smaller models have less headroom to simultaneously perform complex rewriting and strict content preservation. One possible hypothesis is that using the same backbone model both to propose/refine prompts during optimization increase the task difficulty which can favor larger models compared to smaller versions. Taken together, these results support that adaptive prompt optimization is effective across model scales, while larger backbones provide more favorable privacy--utility trade-offs on the hardest tasks which can compete with closed-models.

\begin{table}[h]
    \centering
    \setlength{\tabcolsep}{3pt} 
    \renewcommand{\arraystretch}{1.2}
    
    \resizebox{\columnwidth}{!}{
    \begin{tabular}{lccccc}
    \textbf{Methods}
     & \textbf{DB-Bio} 
     & \textbf{SynthPAI}
     & \textbf{TAB}
     & \textbf{PUPA} 
     & \textbf{MedQA} \\
    \midrule
    \Qwenemoji{}~Qwen3-30B-A3B &  &  &  &  &  \\
    \textit{Seed Prompt} & 64.0/100 & 12.6/92.4 & 36.2/54.0 & 82.3/72.1 & 3.52/58.6 \\
    \textit{Optimized Prompt} & 65.5/100 & 22.5/94.4 & 92.3/56.2 & 98.0/79.3 & 24.6/45.9 \\
    \midrule
    \midrule
    \Qwenemoji{}~Qwen-2.5-7b &  &  &  &  &  \\
    \textit{Seed Prompt} & 41.2/97.3 & 5.88/91.5 & 34.3/55.3 & 79.7/87.3 & 5.18/47.1 \\
    \textit{Optimized Prompt} & 47.1/98.0 & 17.6/86.0 & 90.6/53.2 & 88.4/79.2 & 7.29/41.2 \\
    \bottomrule
    \end{tabular}
    }
    \caption{Performance of Qwen-2.5-7b compared to Qwen3-30B-A3B and their seed prompt version (encoded as \texttt{privacy/utility}).}
    \label{tab:slm_results}
\end{table}

\begin{table*}[t]
\centering
\small
\resizebox{\textwidth}{!}{
\begin{tabular}{lllllc}
\toprule
\textbf{Task} & \textbf{Domain} & \textbf{Privacy Threat} & \textbf{Privacy Metric} & \textbf{Utility Metric} & \textbf{Test Size} \\
\midrule
DB-Bio & Biographies & Identity re-identification & LLM attacker (top-3 accuracy) & Occupation classification & 239 \\
SynthPAI & Social media & Attribute inference & LLM attacker accuracy & ROUGE-1 & 617 \\
TAB & Legal documents & Sensitive span leakage & Span masking recall & Semantic similarity & 555 \\
PUPA & User prompts & PII exposure & PII detection recall & LLM-as-judge quality & 350 \\
MedQA & Medical QA & Patient re-identification & Stylistic similarity & Medical QA & 125 \\
\bottomrule
\end{tabular}
}
\caption{Overview of the five anonymization tasks in our benchmark. Each task presents distinct challenges across domain, threat model, and utility requirements, enabling comprehensive evaluation of adaptive anonymization approaches.}
\label{tab:benchmark_overview}
\end{table*}

\section{Comparison with DP-based Methods}
\label{apx:dp_comparison}

We additionally compare our optimized \Qwenemoji{}~Qwen3-30B-A3B anonymizer against two differentially private rewriting baselines, DP-Prompt and DP-MLM~\citep{meisenbacher-etal-2024-dp}, evaluated at two privacy budgets ($\epsilon \in \{500, 100\}$). Although these methods provide formal DP guarantees and are therefore not directly interchangeable with our empirical privacy objective, the comparison is informative for characterizing the privacy--utility trade-offs induced by DP-based rewriting under the task metrics used in our evaluation. Table~\ref{tab:dp_results} reports privacy/utility results across all five datasets.

Across datasets, DP-based rewriting generally improves privacy as the budget becomes more restrictive, but does so at a severe utility cost. This pattern is especially pronounced on DB-Bio, SynthPAI, and MedQA, where the strongest DP settings achieve substantially higher privacy than our method but collapse utility. On TAB, DP methods reach privacy comparable to or slightly above our model, but still trail it in utility. On PUPA, our optimized model outperforms all DP baselines on both dimensions, achieving 98.0/79.3 versus at best 84.2/49.1 for DP-Prompt. This improvement is likely driven by the subjective nature of the utility evaluation and the targeted assessment of privacy via entity deletion. Overall, these results are consistent with prior work~\citep{meisenbacher-etal-2025-impact}: DP-based methods noise are especially strong to preserve privacy, but often at the expense of substantial utility degradation.

\begin{table}[t]
    \centering
    \setlength{\tabcolsep}{3pt} 
    \renewcommand{\arraystretch}{1.2}
    
    \resizebox{\columnwidth}{!}{
    \begin{tabular}{lccccc}
    \textbf{Methods}
     & \textbf{DB-Bio} 
     & \textbf{SynthPAI}
     & \textbf{TAB}
     & \textbf{PUPA} 
     & \textbf{MedQA} \\
    \midrule
    DP-Prompt~($\epsilon=500$) & 61.2/33.8 & 85.7/5.34 & 95.6/40.4 & 76.7/54.2 & 50.6/11.7 \\
    DP-Prompt~($\epsilon=100$) & 82.1/13.2 & 95.2/1.49 & 99.2/37.4 & 84.2/49.1 & 51.6/9.82 \\
    DP-MLM~($\epsilon=500$)    & 58.1/20.3 & 80.6/2.94 & 91.8/25.0 & 72.9/31.4 & 48.1/7.02 \\
    DP-MLM~($\epsilon=100$)    & 77.2/7.26 & 88.5/0.75 & 95.2/22.4 & 80.0/28.5 & 49.0/5.89 \\
    \midrule
    \rowcolor{SRBG} \Qwenemoji{} (Optimized) & 65.5/100 & 22.5/94.4 & 92.3/56.2 & 98.0/79.3 & 24.6/45.9 \\
    \bottomrule
    \end{tabular}
    }
    \caption{Performance of optimized Qwen3-30B-A3B compared with DP-based rewriting methods across two privacy budgets. Results are reported as privacy/utility, with higher being better on both axes. Smaller $\epsilon$ generally improves privacy but substantially reduces utility.}
    \label{tab:dp_results}
    \vspace{-1em}
\end{table}

\section{Human Evaluation}
\label{apx:human_evalution}

To complement the automatic evaluations, we conducted a small-scale human study on a randomly selected subset of 100 examples from the test split of \textsc{DB-Bio} and \textsc{PUPA}. Seven annotators affiliated with the authors’ organization participated in the study, and each anonymized output was independently evaluated by at least three annotators who were unaware of the underlying model identities. For each example, annotators provided binary judgments across four criteria: \textbf{privacy} (\textit{is the text sufficiently anonymized?}), \textbf{utility} (\textit{is the downstream task still achievable from the anonymized text?}), \textbf{content preservation} (\textit{does the anonymized text preserve the main non-sensitive meaning of the original?}), and \textbf{hallucination avoidance} (\textit{does the anonymized text avoid introducing new facts, details, or events that are not supported by the original text, aside from masking or placeholders?}). We report the mean proportion of positive judgments across annotators. The results are shown in Table~\ref{tab:human_eval}.

Overall, the human evaluation supports the main findings of the automatic experiments. In particular, methods that achieve stronger privacy often do so by sacrificing faithfulness or task usefulness, whereas the optimized model attains one of the best overall balances across criteria. \textsc{OpenPII} preserves utility relatively well and yields strong hallucination avoidance, but its low privacy score indicates that entity-level masking alone is often insufficient to remove identifying information. \textsc{DP-Prompt} reaches high privacy, but collapses on utility and content preservation, suggesting that its anonymizations are frequently over-aggressive and no longer support the intended downstream task. Among the GPT-5-based baselines, Task-Specific Prompting achieves the highest utility and content preservation, but at the cost of weaker privacy, pointing to a more conservative rewriting strategy. \textsc{AF} also preserves utility reasonably well, but obtains by far the lowest hallucination avoidance score, suggesting that iterative adversarial rewriting can introduce unsupported or distorted content. \textsc{RUPTA} achieves strong privacy while maintaining high utility, supported by its specific design for \textsc{DB-Bio}. By contrast, our optimized Qwen model remains close to the top methods on privacy and utility, while improving over \textsc{RUPTA} on content preservation and hallucination avoidance. This suggests that the optimized prompt learns a better compromise between removing sensitive information and preserving the original meaning and factual consistency of the text. Although the study is limited in scale, it provides complementary evidence that the gains observed with LLM-based evaluation also transfer to human judgment.

\begin{table}[t]
\centering
\small
\begin{tabular}{lcccc}
\toprule
\textbf{Method} & \textbf{Priv.} & \textbf{Util.} & \textbf{Cont.} & \textbf{Hall.} \\
\midrule
OpenPII & 0.52 & 0.87 & 0.78 & 0.84 \\
DP-Prompt & 0.90 & 0.08 & 0.27 & 0.78 \\
\midrule
RUPTA (\Openaiemoji{}~GPT-5) & 0.91 & 0.87 & 0.57 & 0.70\\
AF (\Openaiemoji{}~GPT-5) & 0.58 & 0.86 & 0.82 & 0.42 \\
Prompt (\Openaiemoji{}~GPT-5) & 0.62 & 0.91 & 0.86 & 0.80 \\
\midrule
\rowcolor{SRBG} \Qwenemoji{} (Optimized) & 0.89 & 0.85 & 0.63 & 0.81 \\
\bottomrule
\end{tabular}
\caption{Human evaluation of anonymization methods (judgments across \textsc{DB-Bio} and \textsc{PUPA}). Scores are the proportion of positive binary judgments (higher is better) for privacy protection (\textbf{Priv.}), utility preservation (\textbf{Util.}), content faithfulness (\textbf{Cont.}), and absence of hallucinations (\textbf{Hall.}).}
\label{tab:human_eval}
\end{table}

\section{Implementation Details}
\label{apx:implementation_details}

\subsection{Our Framework}
\paragraph{Rich Feedback Generation.}
We generate rich feedback for each task automatically using \Qwenemoji{}~Qwen3-Next-80B-A3B-Instruct, though we observed similar results with other closed and open models. The rich feedback agent transforms simple scalar metrics into detailed, actionable feedback strings that guide the evolutionary prompt optimization process. Figure~\ref{fig:rich_feedback_prompt} shows the instruction provided to the rich feedback agent.

\begin{figure}[!t]
    \centering
    \small
    \fbox{\parbox{\columnwidth}{
    \scriptsize
You are an expert Python developer specializing in the DSPy framework.

I will provide you with a Python class definition used for benchmarking text anonymization. The class typically initializes datasets and `dspy` modules, and defines two evaluation methods: `compute\_utility` and `compute\_privacy`.

Your task is to write the code for a single method: `compute\_overall\_score\_with\_rich\_feedback(self, gold, pred, trace=None, pred\_name=None, pred\_trace=None)`.

This method should implement the following logic:
1.  Execute Metrics: Run the logic found in `compute\_utility` and `compute\_privacy`.
    If the metric relies on a `dspy.ChainOfThought` module or similar, do not just get the final boolean/float result. You must capture the reasoning or specific outputs (e.g., the specific attribute the attacker guessed, the list of missing entities, or the logic the judge used) to provide context.
2.  Compute Overall Score: Calculate the average of the utility and privacy scores (unless the logic implies a different weighting).
3.  Construct Feedback: Build a detailed string (`feedback\_string`) that acts as a prompt for an optimizer. This string must: 
    - State the specific values for Utility and Privacy.
    - Include the qualitative details captured in step 1 (e.g., "The attacker successfully identified [Attribute] because [Reasoning]").
    - Conclude with a one-sentence actionable instruction (e.g., "Try to mask more entities while preserving utility.").
4.  Return Output: Return a `dspy.Prediction` object containing the calculated `score` and the generated `feedback`.

Output only the Python code for this method. }}
    \caption{Instruction for the Rich Feedback Agent. This prompt is used to automatically generate detailed feedback functions from base evaluation metrics.}
    \label{fig:rich_feedback_prompt}
    \vspace{-1em}
\end{figure}

\paragraph{Base vs. Rich Feedback Examples.}
To illustrate the difference between base and rich feedback, consider the TAB dataset. The base feedback simply states: \textit{"The overall score is 0.65, which is the arithmetic mean of the utility score (0.75) and the privacy score (0.55). Try to improve both scores."} In contrast, the rich feedback provides: \textit{"Overall Score: 0.650 (out of 1.0). Score Breakdown: Utility (semantic similarity): 0.750. Privacy (entity masking rate): 0.550. Remaining Sensitive Entities (3): John Smith, 555-1234, john@example.com. Try to mask more entities while preserving the utility of the text."} This additional context enables more targeted instruction refinement.
We present example feedback result implementation in Listing~\ref{code:richfeedbackexample}. The rich feedback includes detailed analysis, specific entities or errors, and actionable guidance. This additional context enables more targeted instruction refinement. 
The remaining tasks implementation is available on our GitHub \url{https://github.com/gabrielloiseau/adaptive-text-anonymization}.
\begin{lstlisting}[language=Python,breaklines=true,showstringspaces=false,caption=Generated rich feedback implementation for TAB., label=code:richfeedbackexample]
def compute_overall_score_with_rich_feedback(
    self, gold, pred, trace=None, 
    pred_name=None, pred_trace=None
):
    # Compute individual metrics
    utility = self.compute_utility(gold, pred, trace)
    privacy = self.compute_privacy(gold, pred, trace)
    overall_score = (utility + privacy) / 2.0

    # Gather remaining entities for context
    remaining_entities = []
    for entity in gold.entity_mentions:
        if (entity["identifier_type"] != "NO_MASK"
            and entity["span_text"] 
                not in pred.anonymized_text):
            remaining_entities.append(
                entity["span_text"])
    remaining_entities = set(remaining_entities)

    # Construct detailed feedback string
    feedback_parts = []
    feedback_parts.append(
        f"Overall Score: {overall_score:.3f}")
    feedback_parts.append("\nScore Breakdown:")
    feedback_parts.append(
        f"  - Utility: {utility:.3f}")
    feedback_parts.append(
        f"  - Privacy: {privacy:.3f}")
    feedback_parts.append(
        f"\nRemaining Sensitive Entities "
        f"({len(remaining_entities)}): "
        f"{', '.join(remaining_entities)}")
    feedback_parts.append(
        "Try to mask more entities while "
        "preserving the utility of the text.")
    feedback_string = "\n".join(feedback_parts)

    return dspy.Prediction(
        score=overall_score, feedback=feedback_string)
\end{lstlisting}

\paragraph{Optimization Hyperparameters.}
All experiments use a fixed rollout budget of $B=1500$ LLM forward passes. During Stage 2 (warm-start), we use early stopping with patience $n=5$ iterations to preserve budget for refinement. In Stage 3, we set the adaptive validation sampling ratio to $\alpha=0.3$, evaluating candidate prompts on 30\% of the validation set in a round-robin fashion. The reflection minibatch size is set to 3 examples across all tasks. We use 111 examples for training and 111 examples for validation, with all remaining data reserved exclusively for testing. The seed prompt is the default instruction generated by DSPy for a minimal anonymization signature (\texttt{text -> anonymized\_text}), without any task-specific engineering.

\subsection{Comparison Methods}

\paragraph{Adversarial Feedback and RUPTA Configuration.}
To ensure fair comparison, we configure both Adversarial Feedback~\cite{staab2025anonymizers} and RUPTA~\cite{yang-etal-2025-robust} with $k=3$ refinement hops, matching the multi-turn interaction budget used in their original implementations. All methods use \Openaiemoji{}~GPT-5-chat as the base model to reflect state-of-the-art closed-source performance. Other available GPT-5 variants did not yield meaningful or consistent improvements on our benchmarks, so we report results using GPT-5-chat for clarity and reproducibility.

\paragraph{Task-Specific Manual Prompts.}
We design task-specific prompts through careful manual engineering, explicitly encoding privacy and utility requirements for each benchmark. These prompts serve as upper bounds on what can be achieved through expert human effort without optimization. The prompts are shown in Figures~\ref{fig:task_specific_tab}--\ref{fig:task_specific_medqa}.

\begin{figure}[!t]
    \centering
    \small
    \fbox{\parbox{\columnwidth}{
    \scriptsize
Given a legal case document in the field `text`, anonymize all personally identifiable information including names, addresses, phone numbers, email addresses, dates of birth, and any other sensitive identifiers. Replace each entity with a generic placeholder (e.g., [PERSON], [LOCATION], [DATE]). Preserve the semantic meaning and legal context of the document. Output the anonymized text in the field `anonymized\_text`. }}
    \caption{Task-specific prompt for TAB.}
    \label{fig:task_specific_tab}
    \vspace{-1em}
\end{figure}

\begin{figure}[!t]
    \centering
    \small
    \fbox{\parbox{\columnwidth}{
    \scriptsize
Given a user query in the field `text`, remove or generalize all personally identifiable information (PII) including names, addresses, phone numbers, email addresses, social security numbers, and credit card numbers. The anonymized query should preserve the user's intent and allow for generating a comparable response. Output the sanitized text in the field `anonymized\_text`. }}
    \caption{Task-specific prompt for PUPA.}
    \label{fig:task_specific_pupa}
    \vspace{-1em}
\end{figure}

\begin{figure}[!t]
    \centering
    \small
    \fbox{\parbox{\columnwidth}{
    \scriptsize
Given a Reddit comment in the field `text`, rewrite it to prevent inference of the author's demographic attributes (age, gender, location, occupation, education, relationship status) while preserving the core content and message. Generalize or remove identifying markers without inventing new information. Output the rewritten comment in the field `anonymized\_text`. }}
    \caption{Task-specific prompt for SynthPAI.}
    \label{fig:task_specific_synthpai}
    \vspace{-1em}
\end{figure}

\begin{figure}[!t]
    \centering
    \small
    \fbox{\parbox{\columnwidth}{
    \scriptsize
Given a celebrity biography in the field `text`, anonymize the text to prevent re-identification of the individual while preserving information about their occupation and achievements. Remove or generalize names, specific locations, organization names, and other identifying details. Maintain the factual content necessary for occupation classification. Output the anonymized biography in the field `anonymized\_text`. }}
    \caption{Task-specific prompt for DB-Bio.}
    \label{fig:task_specific_dbbio}
    \vspace{-1em}
\end{figure}

\begin{figure}[!t]
    \centering
    \small
    \fbox{\parbox{\columnwidth}{
    \scriptsize
Given a medical case description in the field `text`, rewrite it to obfuscate the author's writing style while preserving all clinical information necessary to answer the associated medical question. Vary sentence structure, word choice, and phrasing without altering diagnostic details, symptoms, lab values, or treatment information. Output the rewritten case in the field `anonymized\_text`. }}
    \caption{Task-specific prompt for MedQA.}
    \label{fig:task_specific_medqa}
    \vspace{-1em}
\end{figure}

\section{Computational Costs.}
We report approximate API costs to facilitate reproducibility and cost-aware deployment decisions. For our framework, running the full optimization pipeline locally with open-source models (Mistral-Small, Gemma-3, or Qwen3) incurs costs only for the external evaluation backbone. Using Gemini-2.5-flash for privacy and utility evaluation during optimization and final testing results in approximately \$1 per task per model, covering the 1,500 rollout budget plus validation and test set evaluations. In contrast, the GPT-5-based comparison methods require significantly higher expenditure due to the cost of closed-source inference. Running a single GPT-5-based agent across all test examples costs approximately \$8 per task. These estimates highlight a practical advantage of our approach by shifting the computational burden to locally deployed open-source models and using affordable API-based evaluation only for metric computation. Moreover, even when local GPU resources are unavailable, outsourcing inference for open-source models via trustworthy providers incurs only minimal additional costs due to the competitive pricing of mid-sized models at the time of writing ($< \$0.10$ per task).

\section{Hardware and Code}
We conducted all experiments with two Nvidia Quadro RTX 6000 GPU cards with 24GB memory and Intel Xeon Silver 4114 CPU. The main libraries used include DSPy 3.0.4, HuggingFace transformers 4.57.1, datasets 4.4.1 and sentence-transformers 5.1.2.

\section{Scientific Artifacts}
We used DSPy, TAB, PUPA (MIT), DB-Bio, LUAR (Apache 2.0), SynthPAI (CC-BY-NC-SA-4.0), MedQA (CC-BY-SA 4.0). For models, Mistral-Small and Qwen3 are distributed under Apache 2.0 and Gemma-3 under its dedicated terms of use \url{https://ai.google.dev/gemma/terms}. 

\onecolumn
\section{Task Details}
\label{apx:task_details}
\subsection{DB-Bio}
The DB-Bio task \cite{yang-etal-2025-robust} uses biographical entries from the DBpedia Classes dataset, which contains short, factual descriptions of notable individuals including their profession, achievements, and life events. The privacy objective is to prevent re-identification: an adversarial LLM attempts to infer the identity of the person described in the biography by analyzing residual biographical signals. This represents a realistic threat model where attackers possess extensive world knowledge and reasoning capabilities about celebrities. The utility objective requires preserving sufficient occupational information to maintain accuracy on occupation classification. This task tests the model's ability to remove identifying details (names, specific achievements, temporal markers) while retaining abstract professional characteristics. The tension between identity protection and occupational signal preservation creates a nuanced trade-off that requires careful semantic manipulation.

\begin{trainingexample}{Training Example \,|\, \textsc{DB-Bio}}
\renewcommand{\arraystretch}{1.4}
\begin{tabularx}{\linewidth}{>{\raggedright\arraybackslash}p{2.3cm} X}
  \EXrow{ExpertBG}{Privacy Metric:}{
    Binary re-identification metric: an attacker LLM (Listing~\ref{lst:dbbioprivacy}) proposes the top-3 most likely celebrity names based on the anonymized biography. Privacy score is 1.0 if the true identity is not in the top-3, 0.0 otherwise.
    }
  \EXrow{StateBG}{Utility Metric:}{
    Binary classification accuracy: a classifier LLM (Listing~\ref{lst:dbbioutility}) predicts the person's occupation from the anonymized text.
    }
  \EXrow{IWMBG}{Data Sample: }{
    Geoffrey Harber Diggle (6 December 1902 – 13 February 1993) was a British chess player and writer. Diggle contributed articles to the British Chess Magazine (BCM) from 1933 to 1981, Alexander called Diggle "one of the best writers on chess that I know". In his A Book of Chess, Alexander reproduced in toto Diggle's account, first published in the November and December 1943 BCM, of the de facto 1843 world championship match between Staunton and St. Amant. $[...]$ Diggle later adopted the sobriquet as a pseudonym, writing a series of articles in Newsflash under that name between 1974 and 1986. Chess historian Edward Winter wrote the following in his remembrance of Diggle in CHESS magazine: Specializing in nineteenth-century chess history (particularly the Staunton period), he brought the old masters to life with rare wit and shrewdness. $[...]$
    }
  \EXrow{SRBG}{Optimized Prompt (Mistral-Small):}{
      $[...]$ Anonymize the given text by redacting all personally identifiable information (PII) while ensuring: 1. The occupation or role of the individual(s) remains clear from the context. 2. The text remains useful and coherent for its intended purpose. 3. The anonymized text does not allow identification of the individual(s) through inference or remaining details. $[...]$ Guidelines 1. Redaction Rules - Names: Replace all names (first, last, nicknames, titles) with placeholders (e.g., `[Name]`, `[Nickname]`, `[Lastname]`). - Dates: Replace specific dates, years, or ranges with placeholders. - Locations: Replace cities, states, countries, and specific addresses with placeholders. - Organizations: $[...]$ 2. Preservation Rules - Occupation/Role: Ensure the individual's profession or role is still inferable (e.g., "cyclist," "diplomat," "racing driver").  $[...]$  - Multiple Individuals: Redact all names and identifiers for all mentioned individuals. $[...]$  - Unique Achievements: Redact specific achievements (e.g., "gold medal in 2004 Olympics") to prevent identification. - Scandals/Controversies: Redact details of scandals or controversies that could uniquely identify the individual. $[...]$ Strategy 1. Scan for PII: Identify all names, dates, locations, organizations, and unique achievements. 2. Replace with Placeholders: Use generic placeholders while preserving context. 3. Validate: Check for identifiability and utility.

  }
\end{tabularx}
\end{trainingexample}

\begin{lstlisting}[language=Python,breaklines=true,showstringspaces=false,caption=\centering Privacy attacker signature for DB-Bio., label=lst:dbbioprivacy]
class Privacy(dspy.Signature):
    """You are very knowledgeable about celebrities and
    can identify them from the given anonymized biography. When asked to
    identify individuals, disregard any missing information and context, like
    placeholder '<PERSON>' or 'ORGANIZATION', respond only with the
    names of 3 celebrity candidates most likely being talked by
    the provided anonymized biography text and sort the names from
    highest to lowest probability."""

    biography: str = dspy.InputField()
    celebrity_candidates: list[str] = dspy.OutputField()
\end{lstlisting}
\vspace{-0.1em}
\begin{lstlisting}[language=Python,breaklines=true,showstringspaces=false,caption=\centering Utility evaluator signature for DB-Bio., label=lst:dbbioutility]
class Utility(dspy.Signature):
    """Classify the following biography text into one of the following occupations"""

    biography: str = dspy.InputField()
    occupation: Literal[*self._labels] = dspy.OutputField()
\end{lstlisting}

\subsection{SynthPAI}
SynthPAI~\cite{yukhymenko2024synthetic} addresses the challenging problem of mitigating inference attacks based on weak, distributed profiling signals. The dataset consists of synthetic Reddit-style posts where demographic attributes (age, gender, location) and other personal characteristics can be inferred from writing style, topic choices, and contextual clues rather than explicit mentions. Privacy is evaluated using an adversarial LLM that attempts to infer specific labeled attributes from sanitized text, representing a sophisticated attacker capable of exploiting subtle linguistic patterns and background knowledge. Utility is measured through ROUGE-1 score to ensure the sanitized post retains its core content and readability. This task is particularly challenging because the sensitive information is implicit and diffuse throughout the text rather than localized to specific entities. Effective sanitization requires identifying and mitigating subtle stylistic and contextual cues while preserving the post's communicative value, testing the limits of LLM-based semantic rewriting capabilities.
\begin{trainingexample}{Training Example \,|\, \textsc{SynthPAI}}
\renewcommand{\arraystretch}{1.4}
\begin{tabularx}{\linewidth}{>{\raggedright\arraybackslash}p{2.3cm} X}
  \EXrow{ExpertBG}{Privacy Metric:}{
    Binary re-identification metric: an attacker LLM attempts to infer the target demographic attribute from both original and anonymized text, and a verifier LLM (Listing~\ref{lst:synthpaiprivacy}) checks if the inferences match. Privacy score is 1.0 if the inferences differ (successful anonymization), 0.0 otherwise.
    }
  \EXrow{StateBG}{Utility Metric:}{
    ROUGE-1 F-measure between original and anonymized text.
    }
  \EXrow{IWMBG}{Data Sample (target attribute: Wellington, New Zealand): }{
    Heard rumors 'bout some cheeky taniwha messin' with boaters out on oriental bay – old stories die hard i reckon.
    }
  \EXrow{SRBG}{Optimized Prompt (Gemma~3):}{
      You are an expert at anonymizing text while preserving its meaning and utility. $[...]$ The goal is to remove potentially identifying information about a person, while maintaining the core meaning and sentiment of the original text as much as possible. A high utility is crucial. Here's how to approach the anonymization: 1. Identify potentially identifying information:  Look for details that could reveal someone's occupation, location (city, country), or other private attributes.  This includes: Specific job titles or roles (e.g., "editing suite") References to niche hobbies or cultural practices specific to a region (e.g., "kabuki", "kite battles") – prioritize these for generalization or removal. Details about frequented locations, especially local spots (e.g., "local spots", references to specific festivals). References to financial hardship or specific income levels (e.g. "barely make ends meet"). 2. Anonymize by generalization or removal: Generalization: Replace specific details with broader categories. For example: "kabuki" -> "performance art" "editing suite" -> "workplace" or simply “a work environment” $[...]$ Prioritize retaining the overall message. 3. Preserve Utility:  Strive to maintain the original meaning and intent of the text as much as possible. Your goal is to reduce identifiability, not to rewrite the text. $[...]$ 4. Avoid Over-Anonymization: $[...]$ 5. Context Matters:  Pay attention to the context.  Combine seemingly innocuous details can become identifying if combined. Important Considerations:   Retain conversational elements like "lol" and expressions of emotion unless a specific detail is too revealing. $[...]$
      }
\end{tabularx}
\end{trainingexample}
\begin{lstlisting}[language=Python,breaklines=true,showstringspaces=false,caption=\centering Privacy attacker and verifier signatures for SynthPAI., label=lst:synthpaiprivacy]
class Privacy(dspy.Signature):
    """The following are public comments made on Reddit by one author.
    Your task is to guess the authors target_attribute"""

    comment: str = dspy.InputField()
    target_attribute: str = dspy.InputField()
    guess: str = dspy.OutputField()


class Verif(dspy.Signature):
    """
    You are a helpful assistant that decides whether a prediction and a target refer to the same
    thing or if the prediction is less precise.

    Below I give you several two string pairs, the ground truth and a prediction each, and you
    have to decide whether the prediction is correct. If it is return True. If not write False.
    One correct example is 'usa' and 'United States'. In case the prediction is a less precise
    version of the ground truth, e.g., 'Vancouver' and 'Canada', you should type True.
    For locations and only locations if the prediction is more accurate and also contains the
    ground truth, e.g., prediction is 'London, UK' and ground truth is 'United Kingdom', you
    should type True.
    For occupation unemployed and none should be considered the same.
    """

    string_A: str = dspy.InputField()
    string_B: str = dspy.InputField()
    prediction: bool = dspy.OutputField()
\end{lstlisting}

\subsection{TAB}
The Text Anonymization Benchmark (TAB)~\cite{pilan-etal-2022-text} comprises sensitive text spans extracted from European court case documents. Legal texts contain personal information about parties, witnesses, and case-specific details that must be protected for publication or research use. Privacy is measured through recall of properly masked sensitive spans: the proportion of annotated sensitive information successfully removed or replaced. Utility is assessed via semantic similarity using a cross-encoder model. This task presents a different challenge: rather than full document rewriting, it requires precise identification and localized modification of sensitive spans while maintaining document coherence and legal semantic content. The structured nature of legal text and the availability of gold-standard sensitive span annotations make this task particularly suited for evaluating fine-grained anonymization precision.
\begin{trainingexample}{Training Example \,|\, \textsc{TAB}}
\renewcommand{\arraystretch}{1.4}
\begin{tabularx}{\linewidth}{>{\raggedright\arraybackslash}p{2.3cm} X}
  \EXrow{ExpertBG}{Privacy Metric:}{
    Entity masking recall: the fraction of gold-annotated sensitive spans that no longer appear in the anonymized text.
    }
  \EXrow{StateBG}{Utility Metric:}{
    Semantic similarity using a cross-encoder model (\texttt{cross-encoder/stsb-roberta-base}) that computes similarity scores between original and anonymized documents.
    }
  \EXrow{IWMBG}{Data Sample: }{
    PROCEDURE
    
    The case originated in an application (no. 38007/02) against the Republic of Poland lodged with the Court under Article 34 of the Convention for the Protection of Human Rights and Fundamental Freedoms (“the Convention”) by a Polish national, Mr Artur Warsiński (“the applicant”), on 3 October 2002. The Polish Government (“the Government”) were represented by their Agent, Mr J. Wołąsiewicz of the Ministry of Foreign Affairs. $[...]$ 
    
    THE CIRCUMSTANCES OF THE CASE
    
    The criminal proceedings against the applicant. The applicant, Mr Artur Warsiński, is a Polish national who was born in 1976 and lives in Bytów. The applicant was charged with drug trafficking and detained on remand from 6 April 2002 to 21 May 2004. He was detained in the Słupsk Detention Centre. On 28 May 2003 the Słupsk Regional Court found the applicant guilty of drug trafficking and sentenced him to three years' imprisonment. $[...]$ The envelope also bears a stamp of the Słupsk Regional Court. On 4 December 2002 the applicant sent another letter to the European Court of Human Rights. It was received by the Court on 17 December 2002. It bears a stamp “censored on”, a handwritten date “4 December” and an illegible signature. The envelope also bears a stamp of the Słupsk Regional Court. $[...]$ On 30 June 2003 he received a letter from the President of the Słupsk Regional Court informing him that, according to the information obtained from the Criminal Department of the Słupsk Regional Court, the applicant's correspondence had not been censored. He further stated that the applicant's correspondence had been stamped “censored” by mistake.
    }
  \EXrow{SRBG}{Optimized Prompt (Qwen3):}{
    You are to perform a rigorous, context-aware, and deterministic anonymization of legal or judicial documents using a multi-layered strategy derived from verified best practices and iteratively refined through real-world feedback. The goal is to produce an anonymized\_text output that eliminates **all identifiable entities** while preserving exact structural, legal, factual, procedural, chronological, and linguistic fidelity** of the original. $[...]$ The output is analytically useful for legal, human rights, academic, or policy research contexts without compromising privacy. $[...]$ All replacements must be deterministic and consistent throughout the text: Individuals - Full names → Always `[REDACTED]` if gender unspecified. $[...]$ Legal professionals → Replace with: - `a lawyer practising in [TOWN]`  - `a representative of [INSTITUTION]`  - `a judge at the [CITY] Court of Appeal` - Do **not** use abbreviations (e.g., “R.K.” → `[REDACTED]` only). $[...]$ 1. Scan the document **once**, identifying all entities with precise type (name, date, place, number, etc.). 2. Apply **predefined, reusable replacements** using the exact templates above. $[...]$ 7. **Never** replace named laws, international conventions, or standard legal instruments (e.g., “the Convention”, “Rule 52 § 1”) with placeholders unless the jurisdiction or location enables re-identification. $[...]$ 10. Do not omit any temporal expressions: every date — including in ranges, comparisons, or durations — must be replaced. Output only the fully anonymized text in the `anonymized\_text` field. No explanations.
}
\end{tabularx}
\end{trainingexample}

\subsection{PUPA}
The PUPA task~\cite{zhao2024wildchat} focuses on sanitizing user prompts submitted to conversational AI systems in different languages, using data from real ChatGPT interactions. Users often inadvertently include personally identifiable information (PII) in their prompts—names, addresses, phone numbers, email addresses, and other explicit identifiers. The privacy objective is to detect and remove all labeled PII instances from the text. Utility is evaluated using an LLM-as-a-judge approach: we measure whether responses generated from sanitized prompts maintain comparable quality and relevance to responses from original prompts. This task evaluates the framework's ability to perform entity-level redaction while preserving the semantic intent and context necessary for meaningful AI assistant responses. Unlike the other tasks, PUPA emphasizes explicit identifier removal rather than implicit signal obfuscation, representing a more traditional anonymization scenario adapted to modern LLM interaction contexts.
\begin{trainingexample}{Training Example \,|\, \textsc{PUPA}}
\renewcommand{\arraystretch}{1.4}
\begin{tabularx}{\linewidth}{>{\raggedright\arraybackslash}p{2.3cm} X}
  \EXrow{ExpertBG}{Privacy Metric:}{
    LLM judge (Listing~\ref{lst:pupaprivacy}) that counts PII units remaining in the anonymized query. We rely on an LLM judge following original work \cite{li-etal-2025-papillon} because ground truth labels contains annotated PII which correspond to celebrity names (which in this task is not considered private information).
    }
  \EXrow{StateBG}{Utility Metric:}{
    Generating responses to both original and anonymized queries using \Googleemoji{}~Gemini-2.5-flash, then comparing response quality via an LLM judge (Listing~\ref{lst:pupautility}).
    }
  \EXrow{IWMBG}{Data Sample: }{
    Hi Salar/ Waqas, Well noted. We will arrange for testing once receive the board samples you sent. Would you please also advise the demand and delivery location for measure jig (one free jig per factory location), so that we can send them to you as soon as possible. Thanks for your attention to this matter \begin{CJK*}{UTF8}{gbsn}是什么意思\end{CJK*}
    }
  \EXrow{SRBG}{Optimized Prompt (Qwen3):}{
    You are an assistant tasked with processing text and producing a modified version of it based on specific instructions. Task Description: Given a field called `text`, your goal is to produce a new field called `anonymized\_text`. $[...]$ The tasks can vary widely, from simple redaction to complex rephrasing, translation, or code modification. The type of modification needed is determined by the nature of the input text. $[...]$ PII Redaction/Generalization: Identify and protect PII. This may include, but is not limited to: Names of people and organizations (e.g., "St. Mary\'s Church", "Hermawan DataTables", "Lille Université Club", "Marcq-en-baroeul basketball club"). Replace specific names with generic placeholders (e.g., "Church", "DataTables library", "a university club", "a town basketball club"). $[...]$ Consider generalization (e.g., "aged 18 to 45" can often remain unchanged, but specific birthdates should be removed). Code snippets containing sensitive information (API keys, specific URLs). Any other data that could be used to identify an individual. $[...]$ When the input is code (e.g., CodeIgniter 4 code snippet), maintain the code's functionality. Replace specific library names or project-specific details with generic equivalents. Do not alter the code's behavior unless explicitly requested. $[...]$ Focus on anonymizing project-specific names (e.g. controller/method names) $[...]$ The responses will be evaluated on the level of PII leakage. Aim for a 0\% leakage rate whenever possible. $[...]$
 }
\end{tabularx}
\end{trainingexample}
\begin{lstlisting}[language=Python,breaklines=true,showstringspaces=false,caption=\centering Privacy evaluator signature for PUPA., label=lst:pupaprivacy]
class JudgeLeakage(dspy.Signature):
    """
    Count the number information pieces in `pii` that are simultaneously (i) forms of PII and (ii) are leaked into the prompt given.
    Don't count the cases where the leaked information is just a celebrity name.
    Don't count the cases where the `pii` information does not appear in the prompt.
    """

    pii = dspy.InputField()
    prompt = dspy.InputField()
    num_pii_leaked: int = dspy.OutputField()
\end{lstlisting}
\begin{lstlisting}[language=Python,breaklines=true,showstringspaces=false,caption=\centering Utility evaluator signature for PUPA., label=lst:pupautility]
class JudgeQuality(dspy.Signature):
    """
    You are comparing the quality of two responses, given a user query.
    Is response A at least as good as response B?
    """

    user_query: str = dspy.InputField(desc="The user's request to be fulfilled.")
    response_A: str = dspy.InputField()
    response_B: str = dspy.InputField()
    judgment: bool = dspy.OutputField()
\end{lstlisting}

\subsection{MedQA}
Built on the MedQA dataset~\cite{jin2021disease}, which contains multiple-choice questions from the United States Medical Licensing Examination, this task addresses the sanitization of medical case descriptions containing patient information and clinical details. Privacy is evaluated by measuring the writing style similarity between medical claims extracted from sanitized and original texts. Utility is assessed through the downstream performance of zero-shot medical question answering classifiers applied to sanitized case descriptions. The key challenge is to obfuscate patient-specific details and potentially identifying medical facts while preserving the clinical reasoning structure and diagnostic information necessary for accurate medical inference. This task reflects real-world healthcare data sharing scenarios where regulatory compliance (e.g., HIPAA) demands rigorous privacy protection without compromising clinical utility.
\begin{trainingexample}{Training Example \,|\, \textsc{MedQA}}
\renewcommand{\arraystretch}{1.4}
\begin{tabularx}{\linewidth}{>{\raggedright\arraybackslash}p{2.3cm} X}
  \EXrow{ExpertBG}{Privacy Metric:}{
    Stylometric distance: we compute LUAR embeddings~\cite{rivera-soto-etal-2021-learning} for both original and anonymized case descriptions using the LUAR model, and define privacy as $1 - \text{cosine\_similarity}$. Higher distance indicates successful style obfuscation.
    }
  \EXrow{StateBG}{Utility Metric:}{
    Binary question-answering accuracy: an LLM (Listing~\ref{code:medqautility}) selects the correct diagnosis from multiple-choice options based on the anonymized case description.
    }
  \EXrow{IWMBG}{Data Sample: }{
    A 3-month-old baby died suddenly at night while asleep. His mother noticed that he had died only after she awoke in the morning. No cause of death was determined based on the autopsy. Which of the following precautions could have prevented the death of the baby?
    \begin{flushleft}
    $\bullet$ Placing the infant in a supine position on a firm mattress while sleeping
    
    $\bullet$ Keeping the infant covered and maintaining a high room temperature
    
    $\bullet$ Application of a device to maintain the sleeping position
    
    $\bullet$ Avoiding pacifier use during sleep
    \end{flushleft}
    }
  \EXrow{SRBG}{Optimized Prompt (Mistral-Small):}{
    You are an expert medical text anonymizer. $[...]$ The ultimate goal is to maximize the “Privacy Score” (a measure of writing style dissimilarity – higher is better) without sacrificing the ability to infer the original diagnosis or solution.$[...]$ Preserve Diagnostic/Problem-Solving Information: This is the most important aspect. Do not remove details crucial for diagnosis or answering the question. This includes symptoms, lab values, imaging results, family history, incidence rates, and the context of the presentation. The goal is to make it harder to identify the individual, not harder to arrive at the correct medical conclusion. $[...]$ For questions, ensure the question remains logical and solvable after anonymization. The correct answer should still be obtainable if it was obtainable prior to anonymization. $[...]$ Consider, *cautiously* and *only if it doesn't impact medical accuracy*, replacing a specific biological term with a more general category (e.g., "peripheral artery" -> "blood vessel,"  "lymph node" -> "gland"). $[...]$ Specifically, avoid simply replacing “a 14-year-old boy” with “a patient". Try paraphrasing sentences to achieve a higher “Privacy Score”. $[...]$
}
\end{tabularx}
\end{trainingexample}
\begin{lstlisting}[language=Python,breaklines=true,showstringspaces=false,caption=\centering Implementation Code for the Utility Evaluator., label=code:medqautility]
class MedQAAnswer(dspy.Signature):
    """You are an expert in answering medical exam questions."""

    question: str = dspy.InputField(
        description="The medical question to answer."
    )
    options: list = dspy.InputField(
        desc="List of multiple choice answer options to choose from."
    )
    answer: str = dspy.OutputField(
        description="The correct answer choice. Make sure to only provide the correct answer, and no additional text."
    )
\end{lstlisting}

\begin{table*}[!htb]
\small
\centering
\begin{tabularx}{\textwidth}{lX}
\toprule
Task: \textsc{DB-Bio} & \\
\toprule
Method & Output \\
\midrule
Original & Robin McLaurin Williams (July 21, 1951 - August 11, 2014) was an American stand-up comedian, actor, director, producer, writer, singer and voice artist. Starting as a stand-up comedian in San Francisco and Los Angeles in the mid-1970s, he is credited with leading San Francisco's comedy renaissance. After rising to fame as Mork in Mork \& Mindy (1978-1982), Williams went on to establish a career in both stand-up comedy and feature film acting. He was known for his improvisational skills. \\ 
\midrule
OpenPII & [PERSON\_NAME] [PERSON\_NAME] [PERSON\_NAME] ([DATE] - [DATE]) was an American stand-up comedian, actor, director, producer, writer, singer and voice artist. Starting as a stand-up comedian in [LOCATION\_CITY] [LOCATION\_STATE] and [LOCATION\_CITY] [LOCATION\_STATE] in the [TIME\_PERIOD], he is credited with leading [LOCATION\_CITY] [LOCATION\_STATE] comedy renaissance. After rising to fame as [PERSON\_NAME] in [PERSON\_NAME] \& [PERSON\_NAME] ([YEAR] - [YEAR]), [PERSON\_NAME] went on to establish a career in both stand-up comedy and feature film acting. He was known for his improvisational skills. \\
\midrule
DP-Prompt & A performer with a background in stand-up rose to prominence after early work in clubs and later became famous for a role before establishing a successful career. His work spanned multiple roles, and he became especially known for energetic improvisation. \\
\midrule\midrule
RUPTA (\Openaiemoji{}~GPT-5)&  A person (born and passed away during unspecified times) was a notable figure in the entertainment industry, with roles spanning from various forms of artistry including acting, directing, producing, and voice artistry. Starting as a stand-up comedian in an unspecified location during an unspecified time, this individual is credited with leading an unspecified cultural movement in that location. After gaining recognition in an unspecified role (during an unspecified period), they went on to establish a career in stand-up comedy and feature film acting. \\
\midrule
AF (\Openaiemoji{}~GPT-5) & An individual who was active from the late 20th century into the early 21st century was an entertainer known for work as a comedian, performer, director, producer, writer, singer, and voice artist. Starting as a stand-up comedian in a region of the United States during the 1970s, this person is credited with contributing to a local comedy renaissance. After rising to fame through a popular television role around that time, they went on to establish a career in both stand-up comedy and film acting and became known for their improvisational skills.\\
\midrule
Task-Specific Prompt (\Openaiemoji{}~GPT-5) & [ANONYMIZED PERSON] was an American stand-up comedian, actor, director, producer, writer, singer and voice artist. Starting as a stand-up comedian in San Francisco and Los Angeles in the mid-1970s, [HE/SHE/THEY] is credited with leading San Francisco's comedy renaissance. After rising to fame as a television character in a popular late-1970s sitcom, [HE/SHE/THEY] went on to establish a career in both stand-up comedy and feature film acting. [HE/SHE/THEY] was known for exceptional improvisational skills.\\
\midrule
\pixtralemoji{}~Mistral-Small-3.2-24B & [Name] [Name2] (July [Date1], [Year1] - August [Date2], [Year2]) was an American stand-up comedian, actor, director, producer, writer, singer and voice artist. Starting as a stand-up comedian in [City1] and [City2] in the mid-[Decade], he is credited with leading [City1]'s comedy renaissance. After rising to fame as [Character] in [TV Show] ([Year3]-[Year4]), [Name] went on to establish a career in both stand-up comedy and feature film acting. He was known for his improvisational skills.\\
\midrule
\Googleemoji{}~Gemma-3-27B-it & Born in July 1951 and passing away in August 2014, this individual was a performer working in comedy, acting, directing, and writing. They began their career as a stand-up comedian in California in the mid-1970s and were part of a burgeoning comedy scene in a major West Coast city. After gaining recognition on a popular television sitcom in the late 1970s and early 1980s, they continued to work as a comedian and established a successful career in film. They were particularly well-regarded for their ability to improvise and think on their feet.\\
\midrule
\Qwenemoji{}~Qwen3-30B-A3B & [REDACTED PERSON] ([DATE] - [DATE]) was an [REDACTED NATIONALITY] stand-up comedian, actor, director, producer, writer, singer and voice artist. Starting as a stand-up comedian in [REDACTED LOCATION] and [REDACTED LOCATION] in the mid-[YEAR], [REDACTED PERSON] is credited with leading [REDACTED REGION]'s comedy renaissance. After rising to fame as [NICKNAME] in [REDACTED TV SHOW] ([YEAR]–[YEAR]), [REDACTED PERSON] went on to establish a career in both stand-up comedy and feature film acting. [REDACTED PERSON] was known for [REDACTED PERSON]'s improvisational skills.\\
\bottomrule
\end{tabularx}
\caption{
    Additional qualitative examples for each text anonymization method.
}
\label{tab:ex_apx}
\end{table*}

\end{document}